\documentclass[preprint,12pt]{elsarticle}

\usepackage{amssymb}
\usepackage{amsmath}

\usepackage{geometry}
\usepackage[algo2e]{algorithm2e}
\usepackage{algorithm}
\usepackage{algorithmic}
\usepackage{setspace}

\journal{Neural Networks}

\begin{document}

\begin{frontmatter}



\title{MGCP: A Multi-Grained Correlation based Prediction Network for Multivariate Time Series}


\author{Zhicheng Chen} 
\affiliation{organization={Shenzhen International Graduate School, Tsinghua University},
            city={Shenzhen},
            country={China}}
\ead{czc22@mails.tsinghua.edu.cn}
            
\author{Xi Xiao\corref{cor1}} 
\affiliation{organization={Shenzhen International Graduate School, Tsinghua University},
            city={Shenzhen},
            country={China}}
\ead{xiaox@sz.tsinghua.edu.cn}

\author{Ke Xu}
\affiliation{organization={Tencent AI Lab},
            city={Shenzhen},
            country={China}}
\ead{kaylakxu@tencent.com}

\author{Zhong Zhang}
\affiliation{organization={Tencent AI Lab},
            city={Shenzhen},
            country={China}}
\ead{todzhang@tencent.com}
            
\author{Yu Rong}
\affiliation{organization={Tencent AI Lab},
            city={Shenzhen},
            country={China}}
\ead{royrong@tencent.com}        

\author{Qing Li}
\affiliation{organization={Peng Cheng Laboratory},
            city={Shenzhen},
            country={China}}
\ead{andyliqing@gmail.com}
  
\author{Guojun Gan}
\affiliation{organization={Department of Mathematics, University of Connecticut},
            state={Connecticut},
            country={USA}}
\ead{guojun.gan@uconn.edu}
            
\author{Zhiqiang Xu}
\affiliation{organization={MBZUAI},
            city={Abu Dhabi},
            country={UAE}}
\ead{zhiqiangxu2001@gmail.com}
            
\author{Peilin Zhao\corref{cor1}}
\affiliation{organization={Tencent AI Lab},
            city={Shenzhen},
            country={China}}
\ead{masonzhao@tencent.com}

\cortext[cor1]{Corresponding author}

\begin{abstract}
Multivariate time series prediction is widely used in daily life, which poses significant challenges due to the complex correlations that exist at multi-grained levels.
Unfortunately, the majority of current time series prediction models fail to simultaneously learn the correlations of multivariate time series at multi-grained levels, resulting in suboptimal performance.
To address this, we propose a Multi-Grained Correlations-based Prediction (MGCP) Network, which simultaneously considers the correlations at three granularity levels to enhance prediction performance. Specifically, MGCP utilizes Adaptive Fourier Neural Operators and Graph Convolutional Networks to learn the global spatiotemporal correlations and inter-series correlations, enabling the extraction of potential features from multivariate time series at fine-grained and medium-grained levels.
Additionally, MGCP employs adversarial training with an attention mechanism-based predictor and conditional discriminator to optimize prediction results at coarse-grained level, ensuring high fidelity between the generated forecast results and the actual data distribution. Finally, we compare MGCP with several state-of-the-art time series prediction algorithms on real-world benchmark datasets, and our results demonstrate the generality and effectiveness of the proposed model.
\end{abstract}




\end{frontmatter}



\section{Introduction}
\begin{figure*}[htbp]
\begin{center}
\includegraphics[width=1\textwidth]{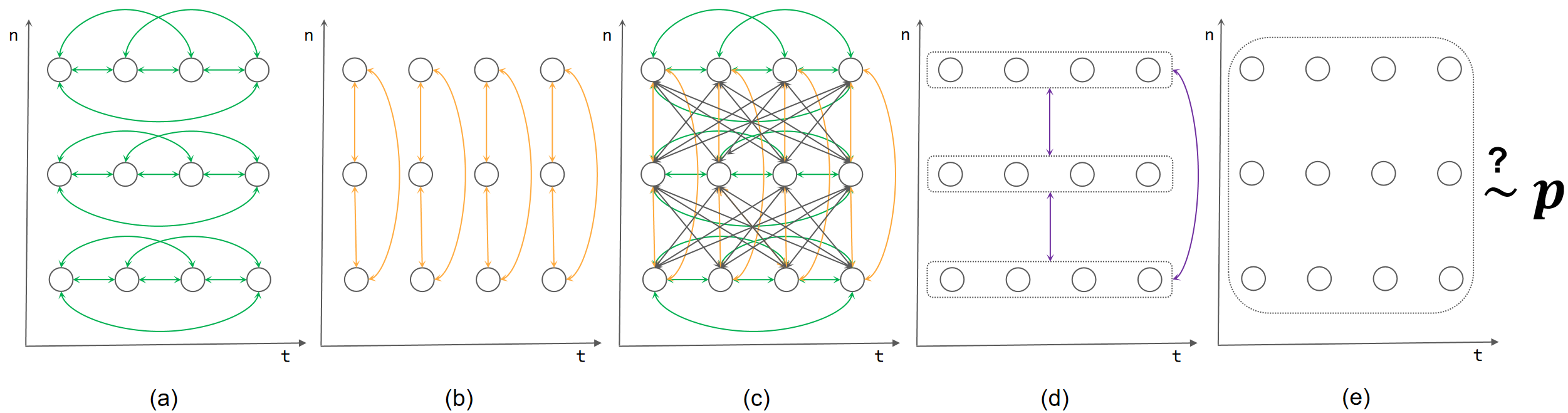}
\end{center}
\caption{Exploring Correlations Patterns of Multivariate Time Series at Multi-Grained Levels: Examples and Illustrations. This figure showcases the correlations of multivariate time series at fine-grained, medium-grained and coarse-grained levels using three time series, each with four timestamps, as examples. At fine-grained level, Panel (a) displays the correlations between different timestamps in each time series, i.e. the temporal correlations, Panel (b) illustrates the correlations between different time series at a certain timestamp, i.e., the cross-sectional correlations and Panel (c) highlights the correlations between arbitrary timestamps for any time series, which we define as the global spatiotemporal correlations. 
At medium-grained level, Panel (d) shows the correlations between different time series, i.e., inter-series correlations.
At coarse-grained level, Panel (e) shows the overall properties of multivariate time series, including whether it follows or approaches a distribution $p$, which may be a conditional distribution in the forecasting task.
}

\label{fig:relationship}
\end{figure*}

Multivariate time series prediction is crucial in various domains that impact our daily lives, including traffic management, financial investment, weather forecasting, and supply chain management. Therefore, it is essential to strive for the highest level of accuracy in multivariate time series prediction. However, with the growing complexity of time series data, accurate time series prediction has become increasingly challenging. Specifically, there are various correlations at multi-grained levels, including fine-grained, medium-grained, and coarse-grained levels.
At fine-grained level, there are several correlations between different timestamps with each time series, including the temporal correlations (Fig. \ref{fig:relationship}(a)), the cross-sectional correlations (Fig. \ref{fig:relationship}(b)), and the global spatiotemporal correlations (Fig. \ref{fig:relationship}(c)). At medium-grained level, a single  time series is treated as a whole to consider its correlations with other time series, primarily the inter-series correlation (Fig. \ref{fig:relationship}(d)). At coarse-grained level, the entire multivariate time series is treated as a whole to consider its correlations with the overall true distribution (Fig. \ref{fig:relationship}(e)). Capturing multi-grained correlations facilitates better multivariate time series forecasting, as considering both local and global features can help the model have better performance\cite{gao2022earthformer,ge2023advancing}.

To address the time series prediction problem, the community initially developed several statistical methods, including ARIMA \cite{liu2016online}, Exponential Smoothing \cite{hyndman2008forecasting} and GARCH \cite{BOLLERSLEV1986307}, which exhibit some favorable statistical properties. However, these models often assume a linear auto-regressive relationship between the time series data and the (high-order) time step differences, which restricts their ability to extract adequate information from the time series. Consequently, these methods may not perform well in handling complex time series data.

In recent years, to tackle complex time series data, many deep learning methods have been proposed, including FC-LSTM \cite{sutskever2014sequence} and SFM \cite{zhang2017stock}, most of which are based on LSTM. However, these simple deep learning methods are not superior to traditional statistical methods \cite{makridakis2018statistical,makridakis2018m4} and are not suitable for long-range time series. To overcome this limitation, advanced deep learning models have been proposed based on convolution, fully connected or attention mechanism, such as N-BEATS \cite{Oreshkin2020N-BEATS:}, TCN \cite{bai2018empirical}, DeepState \cite{rangapuram2018deep}, AST \cite{wu2020adversarial}, Pyraformer \cite{liu2022pyraformer}, FEDformer \cite{zhou2022fedformer}, DLinear \cite{zeng2023transformers}, PatchTST \cite{DBLP:conf/iclr/NieNSK23} and so on. However, most of the above algorithms only capture temporal correlations and do not take into account more complex correlations at fine-grained level, such as the global spatiotemporal correlations, let alone coarser-grained level.

In contrast to them, recent algorithms incorporate Graph Neural Networks (GNN) to enhance multivariate time series prediction, including DCRNN \cite{li2017diffusion}, ST-GCN \cite{yu2017spatio}, GraphWaveNet \cite{wu2019graph}, StemGNN \cite{cao2021spectral}, etc. By utilizing GNN, these methods can effectively capture the inter-series correlations that previous methods could not learn, improving their ability to handle multivariate time series. However, the use of GNN often requires graph structural information, which limits the generality of some of these algorithms. Additionally, most of these methods are limited to learning the correlations at medium-grained level and simple correlations at fine-grained level. Therefore, they cannot capture more complex correlations at fine-grained and coarse-grained levels.

To capture more intricate correlations at fine-grained level, specifically the global spatiotemporal correlations, investigating token mixers \cite{NEURIPS2021_cba0a4ee,NEURIPS2021_4cc05b35,lee-thorp-etal-2022-fnet,NEURIPS2021_07e87c2f,guibas2022efficient} is a promising avenue. Recently, Fourier-based mixers have gained popularity due to their high efficiency and precision in natural language processing and image processing. Specifically, in natural language processing, FNet \cite{lee-thorp-etal-2022-fnet}, which replaces the self-attention sublayer of each Transformer encoder with a Fourier sublayer, has excelled in several tasks and even outperformed the Transformer model in certain cases. In image processing, both Global Filter Networks (GFNs) \cite{NEURIPS2021_07e87c2f} and Adaptive Fourier Neural Operator (AFNO) \cite{guibas2022efficient} have demonstrated strong performance on multiple image classification datasets, surpassing traditional convolutional neural network models in some instances.

However, the aforementioned methods tend to focus primarily on correlations at fine-grained and medium-grained levels, without considering the existence of correlations at coarse-grained level. Building on the recent successes of adversarial learning in image generation and natural language processing \cite{goodfellow2014generative}, adversarial learning has also been applied to various time series-related tasks \cite{mogren2016c,yoon2019time,lin2020using, zhang2023stad}.
These methods have shown good empirical results on real world datasets of various types, because adversarial learning makes the generated data closer to the distribution of the real data.
These methods are proposed for generating new data, whereas time series prediction is to generate the future values of the time series satisfying a conditional distribution with historical values as conditions. Therefore, we can improve adversarial learning by modifying the discriminator into a conditional discriminator, which can let the prediction results approximate the true conditional distribution, that is, learning the correlations at coarse-grained level.

After carefully considering the arguments presented above, our paper introduces an innovative multivariate time series prediction model, called Multi-Grained Correlations-based Prediction (MGCP) Networks, which is capable of extracting correlations at fine-grained, medium-grained and coarse-grained levels. The proposed model leverages Adaptive Fourier Neural Operators (AFNO) and Graph Convolutional Networks (GCN) to dynamically learn the latent space, enabling the feature representations of multivariate time series to capture the global spatiotemporal correlations and the inter-series correlations. 
Additionally, the model utilizes an attention mechanism-based predictor that can efficiently learn long temporal correlations of the time series. 
The proposed model also employs adversarial learning through the conditional discriminator and the predictor to ensure the prediction results approximate the true distribution from the perspective of Wasserstein distance.

Finally, the main contributions of our paper can be summarized as follows:
\begin{itemize}
\item In order to solve the problem of multivariate time series prediction for complex datasets, we propose Multi-Grained Correlations-based Prediction (MGCP) Network. 

\item MGCP can simultaneously consider the correlations at three granularity levels. At fine-grained level, MGCP leverages attention mechanism and AFNO to extract the long-term temporal correlations and the global spatiotemporal correlations, respectively. At medium-grained level, MGCP captures inter-series correlations with GCN. Finally, to capture the correlations at coarse-grained level, MGCP incorporates adversarial learning with an attention mechanism-based predictor and conditional discriminator. 

\item Experimental results on real datasets demonstrate the efficiency and the generality of MGCP. In particular, the ablation experiment proves the rationality of the design of each component of MGCP, and the parameter analysis experiment demonstrates the stability of MGCP.

\end{itemize}

\section{Related Work}
\subsection{Time Series Prediction}
The multivariate time series prediction problem is a rapidly developing area of research. Traditional approaches\cite{liu2016online,hyndman2008forecasting,BOLLERSLEV1986307} have relied on probability and statistics, but more recently, neural networks and deep learning have become popular methods due to their ability to capture complex characteristics of modern time series data. For instance, FC-LSTM \cite{sutskever2014sequence} uses fully-connected LSTM to make prediction, while SMF \cite{zhang2017stock} improves upon the LSTM model. Other models, such as TCN \cite{bai2018empirical}, LSTNet \cite{lai2018modeling}, and DeepGLO \cite{sen2019think}, rely on Convolutional Neural Networks (CNN) to make prediction. N-BEATS \cite{Oreshkin2020N-BEATS:} proposes a model based on forward and backward residual links and a very deep stack of fully connected layers. DeepState \cite{rangapuram2018deep} proposes a model that combines state space models with deep recurrent neural networks. AST \cite{wu2020adversarial} adopts a sparse Transformer to learn a sparse attention map for time series prediction and uses a discriminator to improve the prediction performance at a sequence level. Recently, Pyraformer \cite{liu2022pyraformer} has introduced a pyramidal attention module that uses an inter-scale tree structure to summarize characteristics at different resolutions, while the intra-scale neighboring captures time-dependencies of different ranges. 
FEDformer \cite{zhou2022fedformer} combines the seasonal-trend decomposition method with a frequency enhanced Transformer, effectively capturing both the overarching patterns and intricate details within time series data.
DLinear \cite{zeng2023transformers} is a combination of a decomposition scheme used in FEDformer with linear layers.
PatchTST \cite{DBLP:conf/iclr/NieNSK23} segments time series into subseries-level input tokens for the Transformer and maintains channel-independence by sharing weights across all univariate series.
All of the previously mentioned methods are able to learn the characteristics of individual time series using specific techniques, but they do not learn the correlations between different time series. 

In contrast, there are methods that are designed to learn these correlations. For instance, DCRNN \cite{li2017diffusion} captures spatial dependencies by using bidirectional random walks on a graph and temporal dependencies through an encoder-decoder architecture with scheduled sampling. 
STSGCN \cite{song2020spatial} captures localized spatiotemporal correlations and spatiotemporal data heterogeneity by constructing localized spatiotemporal graphs and using graph convolution operations. 
StemGNN \cite{cao2021spectral} combines Graph Fourier Transform and Discrete Fourier Transform to learn inter-series and temporal correlations. 
Additionally, there are research works that utilize cross-correlation to achieve this goal. For instance, Corrformer \cite{wu2023interpretable} unifies spatial cross-correlation and temporal auto-correlation into a learned multi-scale tree structure, capturing worldwide spatiotemporal correlations.
However, the majority of these methods require predefined graph structures or location information, which limits their generality. Furthermore, these methods are restricted to the temporal correlations and the inter-series correlations, which means they cannot capture more comprehensive correlations at fine-grained and coarse-grained levels. In light of the aforementioned limitations, our objective is to propose an approach that can efficiently consider the correlations of multivariate time series at multi-grained levels.\\
\subsection{Fourier-Based Mixers }
Fourier-based mixers are some techniques that utilize the Fourier transform to mix tokens, which are commonly applied to tackle complex problems in visual and natural language processing.  FNet \cite{lee-thorp-etal-2022-fnet} is a mixer that uses the Fourier transform instead of the self-attention layer in the Transformer, which can effectively solve the computational efficiency and storage space problems of the self-attention layer when processing long sequences. 
However, the mixer does not have learnable parameters, is fixed for token-based mixing, and cannot adapt to specific data distributions.
GFNs \cite{NEURIPS2021_07e87c2f} is a deep learning model for image classification that enables efficient information fusion and feature extraction by learning spatial location-based filter weights on a global scale.
AFNO \cite{guibas2022efficient} improves upon Fourier Neural Operator by enhancing memory and computation efficiency. Specifically, the method optimizes channel mixing weights into block diagonal structures and implements token-adaptive shared weights, as well as sparsifies frequency patterns through soft thresholding and shrinkage. 
\subsection{Generative Adversarial Networks}
Generative Adversarial Networks (GANs) are effective unsupervised generation techniques that have been applied in various domains, including time series related tasks. For instance, C-RNN-GAN \cite{mogren2016c}, TimeGAN \cite{yoon2019time}, DoppelGANger \cite{lin2020using}, and GT-GAN \cite{jeon2022gt} employ GANs to improve time series generation tasks. However, these methods are designed for time series generation rather than forecasting, which is a supervised task. Recently, some studies have utilized adversarial loss as a supplement to the supervision loss in forecasting tasks as a form of regularization or data augmentation. For example, 
AST \cite{wu2020adversarial} employs standard GANs to regularize predictive models and enhance prediction performance at the sequence level. 
TrafficGAN \cite{zhang2019trafficgan} is a traffic flow prediction model based on GAN that utilizes a conditional Wasserstein GAN to improve prediction accuracy by evaluating the difference between generated and real traffic flow sequences.
Nonetheless, the discriminators used in these studies can only determine whether the future values are true or whether the entire sequence is true. To better adapt to multivariate time series prediction tasks, the discriminator should think of the multivariate time series as a whole to optimize the prediction results at the overall distribution level.

\section{Problem Statement and Preliminaries}
\subsection{Problem Statement}
In this part, we first introduce the problem we aim to address. Our input is defined as $\mathcal{G}=(X_{1:T},V_{1:T+\tau})$. Here, $X_{1:T}\in\mathbb{R}^{N\times T}$ denotes the historical observation values across $T$ timestamps, where $N$ is the number of the time series and $T$ is determined by the size of the rolling window. $1:T = [1,\ldots, T]$ is a window within the full time series, and for brevity, we have omitted the starting point of the window. The covariate features (e.g., minute-of-the-hour, hour-of-the-day, etc.) across $T+\tau$ timestamps are represented as $V_{1:T+\tau}\in\mathbb{R}^{N\times (T+\tau)\times d_c}$, where $d_c$ is the number of covariate features and $\tau$ is the size of the predicted steps. The task is to predict $X_{T+1:T+\tau}\in\mathbb{R}^{N\times \tau}$. Assuming that the prediction result of a model $f_\theta$ is $\hat{X}_{T+1:T+\tau}$, i.e.,
\begin{equation}
 \hat{X}_{T+1:T+\tau} = f_\theta(\mathcal{G}).
\end{equation}
The optimization objective of the model can be expressed as:
\begin{equation}
    \mathop{min}\limits_{\theta} \ell\left(X_{T+1:T+\tau},\hat{X}_{T+1:T+\tau} \right),
    \label{Eq:goal}
\end{equation}
where $\ell$ is a commonly used loss function, such as mean squared error.

From a statistical perspective, we can view the problem as finding an optimal approximate distribution that closely approximates the true data distribution. To this end, we define an optimization objective as follows:
\begin{equation}
\mathop{min}\limits_{\hat{p}} \emph{D}\left(p(X_{T+1:T+\tau}|\mathcal{G}),\hat{p}(\hat{X}_{T+1:T+\tau}|\mathcal{G}) \right),\
\label{Eq:goal2}
\end{equation}
where $p$ refers to the true distribution, $\hat{p}$ represents an approximate distribution, and $\emph{D}(\cdot,\cdot)$ is a measure of the difference between the two distributions, such as the Jensen-Shannon divergence or Wasserstein distance.

As a supervised task, time series forecasting often only needs to consider the objective shown in Equation \eqref{Eq:goal}. However, to ensure the algorithm's generalization ability, it is also important to consider the data distribution. Therefore, this article combines Equation \eqref{Eq:goal} and Equation \eqref{Eq:goal2} as our optimization goal to improve prediction performance.
\subsection{Self-Attention and Fourier Neural Operator}
Suppose $Z\in \mathbb{R}^{N\times T\times d}$ is a hidden variable of the multivariate time series, in which $Z[i]\in \mathbb{R}^{T\times d}$ represents the $i$-th time series, $i=1,\cdots,N$. Self-attention can be expressed as: 
\begin{equation}
    \text{Att}(Z[i])=\text{softmax}\left(\frac{Z[i]W_q(Z[i]W_k)^T}{\sqrt{d}}\right)Z[i]W_v,
    \label{Eq:attention}
\end{equation}
where $W_q$, $W_k$, and $W_v$ are weight matrices used to transform the input sequence into query, key, and value vectors.
Through Equation \eqref{Eq:attention}, we can find that self-attention can only learn the correlations between different timestamps of the same time series, and cannot consider the correlations between arbitrary timestamps for any time series. To solve the above problem, we can reshape $Z\in \mathbb{R}^{N\times T\times d}$ into $Z'\in \mathbb{R}^{NT\times d}$ and then use self-attention. Therefore, global self-attention can be express as:
\begin{equation}
    \text{Att}_g(Z)=\text{softmax}\left(\frac{Z'W_q(Z'W_k)^T}{\sqrt{d}}\right)Z'W_v.
    \label{Eq:attention2}
\end{equation}
Equation \eqref{Eq:attention2} indicates that the global spatiotemporal correlations can be taken into account. However, the use of global self-attention, which involves computing attention weights among multiple timestamps of multivariate time series, can significantly increase both time and space complexity. Specifically, the time complexity increases from $O(NTd^2+NT^2d)$ to $O(NTd^2+N^2T^2d)$, and the spatial complexity increases from $O(d^2+NTd+NT^2)$ to $O(d^2+NTd+N^2T^2)$. Therefore, directly utilizing global self-attention is not recommended, and it is imperative to either optimize it or find alternative methods that can achieve equivalent functionality.

In fact, the global self-attention can be viewed as a form of kernel summation, which can be transformed into kernel convolution under certain assumptions. Moreover, due to the convolution theorem, the computation of the global self-attention can be approximated using a Discrete Fourier Neural Operator (Discrete FNO) \cite{li2021fourier}. 
While this transformation requires certain assumptions and involves an approximate operation of converting kernel parameters to learnable parameters, it is undeniable that both methods share the ability to learn global spatiotemporal correlations.

 \textbf{Discrete FNO} \cite{li2021fourier}: The discrete version of Fourier Neural Operator (FNO) \cite{li2021fourier}. Given multivariate time series input $Z\in \mathbb{R}^{N\times T\times d}$, Discrete FNO produces an output $\hat{Z}=\text{FNO}(Z)$, where the learnable parameter is a complex-valued weight matrix denoted as $W\in \mathbb{C}^{N\times T\times d \times d}$. Specifically, for any $(n,t)\in [N]\times[T]$, $\hat{Z}_{n,t}\in \mathbb{R}^{d}$ can be expressed as:
 \begin{equation}
    \left\{
     \begin{aligned}
        &\Bar{Z}_{n,t}=[\text{DFT}(Z)]_{n,t}\\
        &\tilde{Z}_{n,t}=W_{n,t}\Bar{Z}_{n,t}\\
        &\hat{Z}_{n,t}=[\text{IDFT}(\tilde{Z})]_{n,t}
     \end{aligned}
     \right. ,
     \label{Eq:FNO}
 \end{equation}
where DFT and IDFT stand for Discrete Fourier Transform and Inverse Discrete Fourier Transform, respectively. Based on Equation \eqref{Eq:FNO}, it is evident that Discrete AFNO employs distinct channel mixing weights for each timestamp, resulting in significant temporal and spatial complexity, which could cause overfitting issues. However, the strong association between Discrete FNO and global self-attention inspires novel approaches to model the global spatiotemporal correlations, such as utilizing the Fourier mixers.

\textbf{AFNO}\cite{guibas2022efficient}: An improved method designed to mitigate the issue of excessive parameter quantity and overfitting caused by channel mixing in Discrete FNO. The output of AFNO is defined as $\hat{Z}=\text{AFNO}(Z)$, where the learnable parameters are two complex-valued weight matrices $W_1, W_2\in \mathbb{C}^{K\times d/K \times d/K}$ and two complex-valued bias vectors $b_1, b_2\in \mathbb{C}^{K\times d/K}$. Here, $K$ denotes the number of blocks. Specifically, for any $(n,t)\in [N]\times[T]$, the expression for $\hat{Z}_{n,t}\in \mathbb{R}^{d}$ is:
     \begin{equation}
    \left\{
     \begin{aligned}
        &\Bar{Z}_{n,t}=[\text{DFT}(Z)]_{n,t}\\
        &\Bar{Z}_{n,t}^{(1:K)}=\text{Reshape}(\Bar{Z}_{n,t},K,d/K)\\ &\tilde{Z}_{n,t}^{(k)}=S_a\left(W_2^{(k)}\sigma(W_1^{(k)}\Bar{Z}_{n,t}^{(k)}+b_1^{(k)})+b_2^{(k)}\right)\\
        &\tilde{Z}_{n,t}=\text{Concat}(\tilde{Z}_{n,t}^{(1:K)})\\
        &\hat{Z}_{n,t}=[\text{IDFT}(\tilde{Z})]_{n,t}
     \end{aligned}
     \right. ,
 \end{equation}
where $k=1,\cdots, K$, $\text{Reshape}(\cdot,K,d/K)$ means reshaping a variable in dimension $d$ into a variable of dimension $K\times d/K$, $S_a(x)=\text{sgn}(x)\text{max}\{|x|-a,0 \}$ is used for soft thresholds and contraction, and $a$ is a tuning parameter that controls the sparsity.
The time complexity of AFNO is $O(NTd^2/K+NTd \log NT)$ and the spatial complexity is $O(NTd+d^2/K)$. This method is obviously optimized in time and space, and cross-token adaptive sharing of weights is more generalized than Discrete FNO. 

\textbf{Masked AFNO:} A method that adds a masking mechanism to AFNO. In our work, we use $X_{T+1:T+\tau}$ in the training phase, and it is replaced by $\textbf{0}$ or noise during the testing phase. Therefore, In order not to affect the model, we need to ensure that $X_{T+1:T+\tau}$ does not affect $X_{1:T}$ when using AFNO, so we need to extend AFNO slightly. Suppose $Z_{1:T+\tau} \in \mathbb{R}^{N\times (T+\tau)\times d}$ is a hidden variable of $X_{1:T+\tau}$, and define the output of Masked AFNO is
 $\hat{Z}_{1:T+\tau}=\text{AFNO}_{\mathcal{M}}(Z_{1:T+\tau})$. 
 $\text{AFNO}_1(\cdot)$ and $\text{AFNO}_2(\cdot)$ are two AFNO networks that share parameters. Therefore, $\hat{Z}_{1:T+\tau}$ can be expressed as:
\begin{equation}
    \left\{
     \begin{aligned}
        &Z^{'}_{1:T}=\text{AFNO}_1(Z_{1:T})\\
        &Z^{''}_{1:T+\tau}=\text{AFNO}_2(Z_{1:T+\tau})\\
        &\hat{Z}_{1:T+\tau}=\text{Concat}(Z^{'}_{1:T},Z^{''}_{T+1:T+\tau})
     \end{aligned}
     \right. .
 \end{equation}
 Masked AFNO can ensure that the historical value will not be affected by the future predicted value, but also does not affect the future forecast value to learn the global spatiotemporal correlations.

\section{Methods}

\begin{figure}[htbp]
\begin{center}
\includegraphics[width=0.36\textwidth]{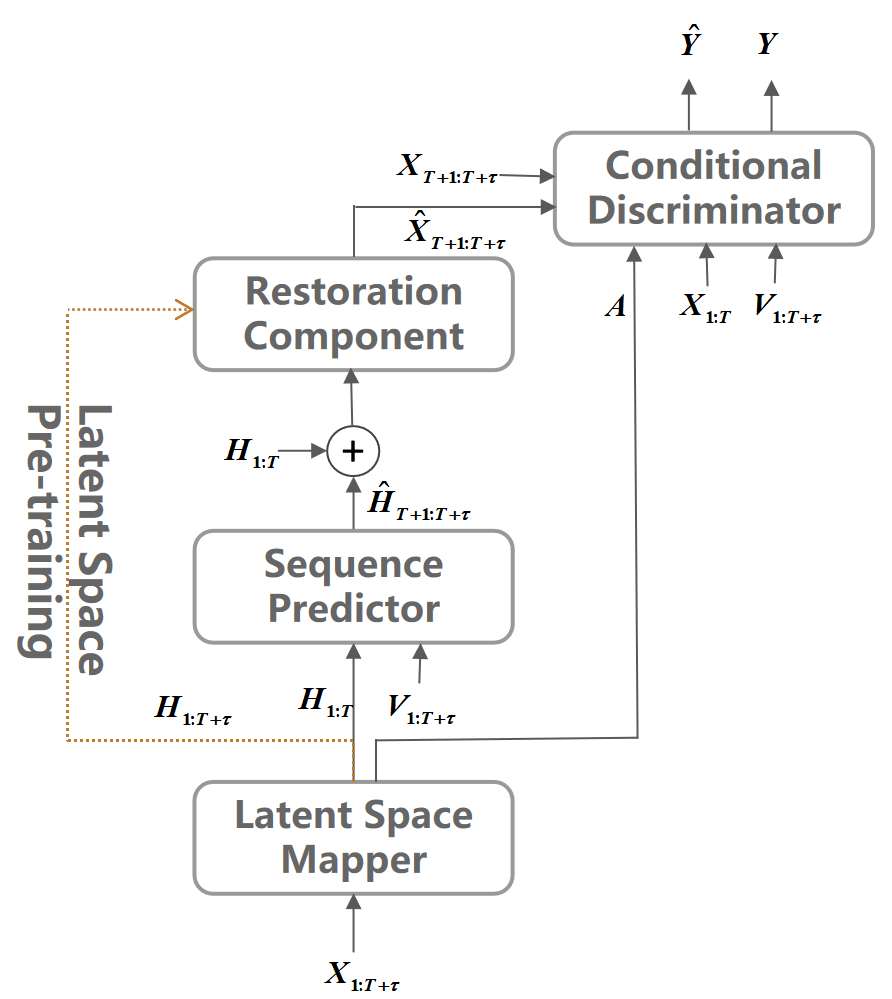}
\end{center}
\caption{Workflow of Multi-Grained Correlations-based Prediction Network. The latent space mapper mainly includes Masked AFNO and GCN, as shown in Fig.~\ref{fig:mapper}. Both the sequence predictor and the conditional discriminator is composed of an attention-based encoder and  an attention-based decoder. The restoration component consists of the AFNO and multi-layer fully connected networks.
}

\label{fig:MGCP}
\end{figure}

In this section, we propose a time series prediction model, named as Multi-Grained Correlations-based Prediction (MGCP) Network, which is capable of simultaneously learning the correlations of multivariate time series at fine-grained, medium-grained and coarse-grained levels.
Specifically, MGCP contains four components: the latent space mapper, the sequence predictor, the restoration component, the conditional discriminator, and its workflow is shown in Fig. \ref{fig:MGCP}. The latent space mapper maps the time series features to the latent space, the sequence predictor makes prediction in the latent space, the restoration component restores the prediction results back to real data space, and the conditional discriminator conducts adversarial learning on the prediction results. 

These four components will be explained in detail in the next three subsections.

\subsection{The Latent Space Mapper }

\begin{figure}[htbp]
\begin{center}
\includegraphics[width=0.23\textwidth]{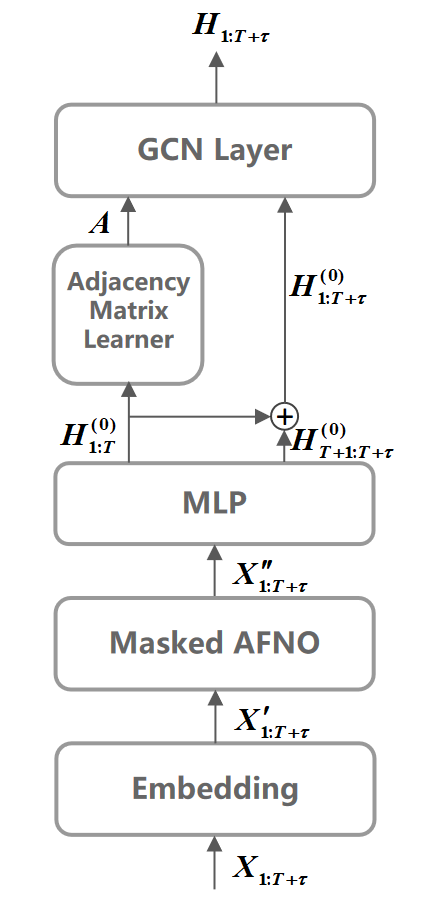}
\end{center}
\caption{The structure of the latent space mapper.}
\label{fig:mapper}
\end{figure}

Inspired by \cite{yoon2019time}, we establish a latent space mapper that enables the full extraction of feature representations from multivariate time series at both fine-grained and medium-grained levels. Specifically, the latent space mapper can map multivariate time series features to a latent space where the global spatiotemporal correlations and the inter-series correlations of multivariate time series can be considered. The structure of the latent space mapper is shown in Fig. \ref{fig:mapper}, and it can be expressed as:  
\begin{equation}
    H=\Phi_{lsm}(X_{1:T+\tau};\theta_{m}),
    \label{Eq:lsm}
\end{equation}
where $\theta_{m}$ is the parameter of the latent space mapper, $X_{1:T+\tau} \in \mathbb{R}^{N \times (T+\tau) \times 1}$ is the input, $H \in \mathbb{R}^{N \times (T+\tau) \times d_l}$ is the output, and $d_l$ is the dimension of the latent space.
$\Phi_{lsm}$ consists of two core parts: Masked AFNO and GCN, which are used to learn the global spatiotemporal correlations at fine-grained level and the inter-series correlations at medium-grained level, respectively. 
Specifically, the calculation of $\Phi_{lsm}$ involves a sequence of steps, which can be summarized by the following equation:
\begin{equation}
    \left\{
        \begin{aligned}
            &X_{1:T+\tau}^{'}=\text{Emb}(X_{1:T+\tau})\\
            &X_{1:T+\tau}^{''}= \text{AFNO}_{\mathcal{M}}(X_{1:T+\tau}^{'})\\
            &H^{(0)} =\text{MLP}(X_{1:T+\tau}^{''})\\
           &H^{(l)}=		\sigma\left(D^{-\frac{1}{2}} A D^{-\frac{1}{2}} H^{(l-1)}\right)
        \end{aligned}
    \right. ,
\label{Eq:mapper}
\end{equation}
where Emb($\cdot$) consists of fully connected networks and positional encoding operations, $l=1, \cdots, L$ represents the layer number, $L$ is the total number of graph convolution, $H=H^{(L)}$ is final output of the latent space mapper, $\sigma$ is activation function, and $A$ is the potential adjacency matrix.

Several studies on traffic flow series prediction \cite{li2017diffusion, song2020spatial} employ a distance-based method to construct the adjacency matrix. However, this approach may not be applicable to all datasets.
To address this limitation, we refer to the method proposed in \cite{cao2021spectral}, which utilizes the feature representations of multivariate time series to learn the latent adjacency matrix. 
The calculation process of potential adjacency matrix $A$ can be expressed as:
\begin{equation}
	\left \{
	\begin{aligned}
		M&=\text{GRU}(H_{1:T}^{(0)})\\
		\tilde{A}&=\text{softmax}(\frac{MW_1^{'} (MW_2^{'})^T}{\sqrt{d_k}})\\
		A&=\frac{1-\beta}{2}(\tilde{A}+\tilde{A}^T)+\beta I_N
	\end{aligned}
	\right. ,
	\label{Eq:adj}
\end{equation}
where GRU is Gated Recurrent Unit, $M\in \mathbb{R}^{N\times d_m}$ is intermediate variable, $W_1^{'},W_2^{'}\in \mathbb{R}^{d_m\times d_k}$ is a learnable weight matrix, $d_m$ and $d_k$ are the sizes of the corresponding dimensions, and $\beta \in [0,1]$ is the weight of the identity matrix in the adjacency matrix.


By using Masked AFNO and GCN, the feature representations from multivariate time series data at fine-grained and medium-grained levels can be effectively learned, allowing subsequent components to efficiently extract features and make accurate prediction. Additionally, the adjacency matrix used in the graph convolution operation is obtained directly from the time series features, which ensures the generality of the network.

\subsection{The Sequence Predictor and The Restoration Component}
The goal of the sequence predictor is to predict the future values based on the historical values in the latent space and the covariate features, and it can be expressed as:
\begin{equation}
    \hat{H}_{T+1:T+\tau}=\Phi_{sp}(H_{1:T},V_{1:T+\tau};\theta_{p}),
    \label{Eq:sp}
\end{equation}
where $\theta_{p}$ is the parameter of the sequence predictor, $\hat{H}_{T+1:T+\tau} \in \mathbb{R}^{N \times \tau \times d_l}$ is the prediction result in the latent space.

Various options are available for predictor, such as LSTM, but in our study, we utilize the Transformer's encoder-decoder architecture \cite{vaswani2017attention} as the predictor to ensure its ability to the learn long-term temporal correlations. Therefore, we implement $\Phi_{sp}$ as:
\begin{equation}
    \left\{
        \begin{aligned}  
       &E_f=\text{Encoder}([H_{1:T},V_{1:T}]) \\
       &\hat{H}_{T+1:T+\tau}=\text{Decoder}(E_f,V_{T+1:T+\tau})
            \label{Eq:predictor}
        \end{aligned}
    \right. ,
\end{equation}
where $\text{Encoder}$ and $\text{Decoder}$ both consist of attention mechanism and fully connected networks. To simplify notation, we use the symbol $[\cdot, \cdot]$ to denote the concatenation of two variables along the last dimension, and this convention applies to subsequent statements as well.

Through the action of the sequence predictor, the prediction result $\hat{H}_{T+1:T+\tau} \in \mathbb{R}^{N \times \tau \times d_l}$ in the latent space can be obtained. Then the restoration component remaps the prediction result in the latent space back to the real data space to achieve the final prediction. Let $\hat{H}=\text{Concat}(H_{1:T},\hat{H}_{T+1:T+\tau})$, and the restoration component can be expressed as:
\begin{equation}
\hat{X}=\Phi_{rc}(\hat{H};\theta_r),
    \label{Eq:restoration}
\end{equation}
where $\theta_{r}$ is the parameter of the restoration component, $\hat{X}_{T+1:T+\tau}\in \mathbb{R}^{N \times \tau \times 1}$ is the final prediction. 
 The structure of $\Phi_{rc}$ is similar to that of the latent space mapper, except that $\Phi_{rc}$ uses $\text{AFNO}(\cdot)$ instead of $\text{AFNO}_{\mathcal{M}}(\cdot)$, and does not include graph convolution operations.

\subsection{The Conditional Discriminator}
To learn the correlations at coarse-grained level, we employ WGAN-gp\cite{gulrajani2017improved} to optimize the conditional Wasserstein distance between the prediction results and the real data. Therefore, we design a conditional discriminator $\Phi_{cd}$ to calculate the conditional Wasserstein distance, which can be expressed as:
\begin{equation}
    \left\{
        \begin{aligned}
            &Y =\Phi_{cd}(X_{T+1:T+\tau}|\mathcal{G};\theta_{d}) \\
            &\hat{Y} =\Phi_{cd}(\hat{X}_{T+1:T+\tau}|\mathcal{G};\theta_{d})
        \end{aligned}
    \right.,
\label{Eq:cd}
\end{equation}
where $\theta_{d}$ is the parameter of $\Phi_{cd}$, $X_{T+1:T+\tau}$ and $\hat{X}_{T+1:T+\tau}$ are the two inputs,  $\mathcal{G}=(X_{1:T},V_{1:T+\tau})$ is the condition. Assuming that the conditional discriminator has been sufficiently trained, the Wasserstein distance between $p(X_{T+1:T+\tau}|\mathcal{G})$ and $\hat{p}(\hat{X}_{T+1:T+\tau}|\mathcal{G})$ can be calculated as:
\begin{equation}
\begin{aligned}
    &\sup_{|{\Phi_{cd}}|_L\leq 1}\left\{ \underset{X_{\tiny T+1:T+\tau}|\mathcal{G} \sim p}{\mathbb{E}}[Y] - \underset{\hat{X}_{T+1:T+\tau}|\mathcal{G} \sim \hat{p}}{\mathbb{E}}[\hat{Y}] 
    \right\} ,
\end{aligned}
\end{equation}
where $|{\Phi_{cd}}|_L\leq 1$ indicates that $\Phi_{cd}$ satisfies the Lipschitz continuity condition.
Following \cite{gulrajani2017improved}, the loss of the conditional discriminator can be expressed as:
\begin{equation}
\begin{aligned}
    \mathcal{L}_D &=\hat{Y}-Y+\alpha\mathcal{L}_{gp},
\end{aligned}
\label{Eq:discriminator loss}
\end{equation}
where $\mathcal{L}_{gp}$ is gradient penalty used to enforce the Lipschitz continuity condition of the conditional discriminator
and the hyperparameter $\alpha$ is used to adjust the weight of the gradient penalty. 
$\mathcal{L}_{gp}$ can be expressed as:
\begin{equation}
    \left\{
    \begin{aligned}
    &\Bar{X}_{T+1:T+\tau}=\epsilon X_{T+1:T+\tau}+(1-\epsilon)\hat{X}_{T+1:T+\tau}\\
    &\mathcal{L}_{gp}=\left(\Vert\nabla_{\Bar{X}_{T+1:T+\tau}}\Phi_{cd}(\Bar{X}_{T+1:T+\tau}|\mathcal{G};\theta_{d})\Vert-1\right)^2
    \label{Eq:gradient penalty}
    \end{aligned}
    \right. ,
\end{equation}
where $\epsilon\in \mathbb{U}(0,1)$ is a random number and $\Vert\cdot\Vert$ represents the Euclidean norm. Specifically, we use graph convolution operations and an attention encoder-decoder structure to construct $\Phi_{cd}$. First, we use the adjacency matrix $A$ calculated from Equation \eqref{Eq:adj} to perform graph convolution operations on multivariate time series. Then, the encoder encodes $X_{1:T}$ and $V_{1:T}$, the decoder uses the encoding result and $V_{T+1:T+\tau}$ to decode $X_{T+1:T+\tau}$ or $\hat{X}_{T+1:T+\tau}$, and finally uses MLP to map the decoding result to obtain $Y\in\mathbb{R}$ or $\hat{Y}\in\mathbb{R}$. Let $\Bar{A}=D^{-\frac{1}{2}} A D^{-\frac{1}{2}}$, so $\Phi_{cd}$ can be implemented as:
\begin{equation}
    \left\{
        \begin{aligned}  
       &E_f^{'}=\text{Encoder}([\Bar{A}X_{1:T},V_{1:T}]) \\
       &Y=\text{MLP}(\text{Decoder}([\Bar{A}X_{T+1:T+\tau},V_{T+1:T+\tau}],E_f^{'}))\\
       &\hat{Y}=\text{MLP}(\text{Decoder}([\Bar{A}\hat{X}_{T+1:T+\tau},V_{T+1:T+\tau}],E_f^{'}))
        \label{Eq:discriminator}
        \end{aligned}
    \right. .
\end{equation}

With the conditional discriminator, MGCP can make the distribution where the prediction results are located closer to the true distribution through adversarial learning.

\subsection{Training and Testing Process}
\textbf{Training Process:}
The training process of MGCP model is divided into two steps, including the latent space pre-training and adversarial training.

In the first step, we exclusively train the latent space mapper and the restoration component with the objective of minimizing the reconstruction loss of the transformed time series. The loss function of the first step is defined as:
\begin{equation}
    \mathcal{L}_R =\ell(X_{1:T+\tau},\tilde{X}_{1:T+\tau}),
\label{Eq:train1}
\end{equation}
where $\tilde{X}=\Phi_{rc}(H;\theta_r)$, $\ell$ is a commonly used loss function, such as mean squared error.


In the second step, we jointly train the four components, and while the predictor and the conditional discriminator learn adversarially, we also continue to dynamically learn the latent space. In general, we optimize the predictor training loss shown in Equation \eqref{Eq:predictor training loss} and adversarial loss shown in Equation \eqref{Eq:discriminator loss} sequentially.
\begin{equation}
\begin{small}
        \begin{aligned}
            \mathcal{L}_P &=\ell(X_{T+1:T+\tau},\hat{X}_{T+1:T+\tau})+\lambda \mathcal{L}_R+ \gamma log(1-\hat{Y}) ,\\
        \end{aligned}
\label{Eq:predictor training loss}
\end{small}
\end{equation}
where $\mathcal{L}_P$ is the predictor training loss, including three parts: prediction loss, reconstruction loss, and adversarial loss. $\lambda$ and $\gamma$ are hyperparameters used to adjust the weights of different parts. 
\begin{algorithm}[ht]
\setstretch{1.2}
  \caption{ The Training Procedure of The Multi-Grained Correlations-based Prediction Network.}
  \label{alg:MGCP}
\begin{algorithmic}
 \STATE{\bfseries Input:} Set trade-off hyperparameters $\lambda$,  $\gamma$,  $\beta$, $\alpha$, total learning rate $\eta$, and discriminator learning rate $\eta_d$.
\\
  \STATE {\bfseries First step:} Pre-training the latent space. 
  \REPEAT
  \STATE Randomly sample $X_{1:T+\tau}$ from training dataset.
  \STATE$ (\theta_m, \theta_r) =  (\theta_m, \theta_r)  - \eta \nabla_{\theta_m,\theta_r} \mathcal{L}_R$
  \UNTIL{reach pre-training times.}
\end{algorithmic}

\begin{algorithmic}
  \STATE {\bfseries Second step:} Adversarial training.

  \REPEAT
  \STATE Randomly sample $X_{1:T+\tau}$, $V_{1:T+\tau}$ from training dataset.
  \STATE$(\theta_m,\theta_r,\theta_p,\theta_d)= (\theta_m,\theta_r,\theta_p,\theta_d) - \eta \nabla_{\theta_m,\theta_r,\theta_p,\theta_d} \mathcal{L}_P$
  \STATE$\theta_d = \theta_d  - \eta_d \nabla_{\theta_d} \mathcal{L}_D$ 
  \UNTIL{reach adversarial training times.}
\end{algorithmic}
\end{algorithm}
Finally, Algorithm $1$ summarizes the training procedure of our proposed Multi-Grained Correlations-based Prediction (MGCP) Network.

\textbf{Testing Process}
During the testing process, we utilize only three components, excluding the conditional discriminator, to generate prediction results. What's more,
unlike the training process, $X_{T+1:T+\tau}$ is not available during the testing process. Therefore, when testing, we feed $\text{Concat}(X_{1:T},\textbf{0})$ instead of $X_{1:T+\tau}$ into MGCP, where $\textbf{0}\in \mathbb{R}^{N \times \tau}$ is an all-zero matrix. Instead of $\textbf{0}$, other options are also possible, such as the mean of $X_i$ for all $i\in\{1,\ldots,T\}$.

\section{Experiments} 
\label{sec:experiments}
In this section, we firstly evaluate our proposed MGCP by comparing it with some state-of-the-art baseline methods. Subsequently, we conduct an ablation study and parametric analysis of MGCP to investigate the effects of its individual components, and examine the impact of hyperparameters on the performance of MGCP. Finally, we will analyze the performance of MGCP over longer forecast steps, and compare the effects of different mixers on both performance and cost.

\subsection{Datasets and Setup}
\begin{table}[t]
\centering
\caption{Descriptions of all datasets. }
\renewcommand\arraystretch{1.2}
\tabcolsep=0.12cm
\begin{tabular}{l|cccc}
\hline
dataset  &\#nodes &\#timesteps   &frequency  & start time  \\
\hline
METR-LA   &207 &34272 &5min  &03/01/2012     \\
PEMS-BAY  &325 &52116 &5min  &01/01/2017       \\
PEMS03    &358 &26209 &5min  &09/01/2018       \\
PEMS04    &307 &16992 &5min  &01/01/2018       \\
PEMS07    &228 &12672 &5min  &05/01/2017    \\
PEMS08    &170 &17856 &5min  &07/01/2016      \\
Solar     &137 &52560 &10min  &01/01/2006      \\
Electricity &321 &26304 &1hour  &01/01/2012   \\
ECG5000   &140 &5000  &-  &-     \\
\hline
\end{tabular}
\label{tab:dataset}
\end{table}

To evaluate the performance of MGCP, we conducted experiments on a total of six traffic datasets, including PEMS-BAY, PEMS03, PEMS04, PEMS07, PEMS08 \cite{chen2001freeway} and METR-LA \cite{jagadish2014big}. Additionally, we evaluated MGCP on three other datasets, including the Solar \cite{lai2018modeling}, Electricity \cite{asuncion2007uci} and ECG5000 \cite{chen2015ucr} datasets. Table \ref{tab:dataset} provides a detailed description of each dataset.

We take the same train-valid-test split as reported by \cite{cao2021spectral} (i.e. 7:2:1 on datasets METR-LA, PEMS-BAY, PEMS07, Solar, Electricity, ECG5000 and 6:2:2 on datasets PEMS03, PEMS04, PEMS08). The input of proposed model is $\mathcal{G}=(X_{1:T},V_{1:T+\tau})$, and we utilize the hour-of-the-day feature and the day-of-the-month feature to structure covariate feature $V_{1:T+\tau}$ on the Electricity  dataset, use zero matrix (or noise matrix) to structure it on the ECG5000 dataset, and use the minute-of-the-hour feature and the hour-of-the-day feature to structure it on the remaining datasets. We use the Adam optimizer to train the model for 50 epochs in total, with a total learning rate of $0.001$ and a discriminator learning rate of $0.0005$. In the following experiments, we use Mean Absolute Errors (MAE), Mean Absolute Percentage Errors (MAPE) and Root Mean Squared Errors (RMSE) to measure the performance of all the algorithms. 

\subsection{Baselines}
\label{sec:baseline}
To evaluate the performance of MGCP, we compare it with several state-of-the-art baselines, including:
\begin{itemize}
\item LSTM-based models, including FC-LSTM \cite{sutskever2014sequence} and SFM \cite{zhang2017stock}.
\item   A number of CNN-based models, including TCN \cite{bai2018empirical}, LSTNet \cite{lai2018modeling} and DeepGLO \cite{sen2019think}.
\item Attention-based models, including AST \cite{wu2020adversarial}, Pyraformer \cite{liu2022pyraformer}, FEDformer \cite{zhou2022fedformer}, PatchTST \cite{DBLP:conf/iclr/NieNSK23} and Corrformer\cite{wu2023interpretable}.
\item Some GNN-based models, including DCRNN \cite{li2017diffusion}, ST-GCN \cite{yu2017spatio}, GraphWaveNet \cite{wu2019graph} and StemGNN \cite{cao2021spectral}. 
\item Several models for other special mechanisms (i.e. fully connected network, state space model), including N-BEATS \cite{Oreshkin2020N-BEATS:}, DeepState \cite{rangapuram2018deep} and DLinear \cite{zeng2023transformers}.
\end{itemize}
For these baselines, most of their performance comes from \cite{cao2021spectral}, and the rest are implemented according to the code provided by the corresponding paper. Some GNN-based models, such as DCRNN and ST-GCN, can only be executed on datasets with graph structure information. Consequently, they are not applicable to the Solar, Electricity, and ECG5000 datasets. Moreover, the use of Corrformer necessitates location information, which restricts its performance evaluation to the METR-LA, PEMS-BAY, and PEMS07 datasets that provide this information.

{\bf Implementation.} The source codes of our proposed method and the corresponding datasets will  be publicly available, along with the camera-ready version of the sub-mission if accepted.

\subsection{Comparison of Results}
Table \ref{tab:experiments} summarizes the performance of all the compared algorithms over the traffic datasets, and the results of the remaining datasets are summarized in Table \ref{tab:experiments2}, which shows only some of the newer and better performing methods due to space constraints. From these results, we can make several observations.

First of all, we notice that ARIMA is mediocre on most datasets. This is because traditional statistical methods have difficulty learning complex relationships within complex data.

\begin{table*}
\centering
\begin{footnotesize}
\caption{ Results of all methods for traffic datasets}
\renewcommand\arraystretch{1.1}
\tabcolsep=0.1cm
\begin{tabular}{lccc|ccc|ccc}
\hline
Datasets &       &METR-LA     &  &      & PEMS-BAY  & &      & PEMS07     & \\
\hline
Metrics &\text{  }MAE\text{  } &\text{  }RMSE\text{ }  &MAPE(\%) &\text{  }MAE\text{  } &\text{  }RMSE\text{ }  &MAPE(\%)&\text{  }MAE\text{  } &\text{  }RMSE\text{ }  &MAPE(\%)
     \\
\hline
ARIMA &3.57 &8.00 &8.26 &1.50 & 3.33 &3.10 &2.25 &4.29 &5.08\\
FC-LSTM   &3.44 &6.3 &9.6 &2.05 &4.19 &4.8 &3.57 &6.2 &8.6   \\
SFM &3.21 &6.2 &8.7 &2.03 &4.09 &4.4 &2.75 &4.32 &6.6 \\
N-BEATS &3.15 &6.12 &7.5 &1.75 &4.03 &4.1 &3.41 &5.52 &7.65 \\
DCRNN &2.77 &5.38 &7.3 &1.38 &2.95 &2.9 &2.25 &4.04 &5.30 \\
LSTNet &3.03 &5.91 &7.67 &1.86 &3.91 &3.1 &2.34 &4.26 &5.41 \\
ST-GCN &2.88 &5.74 &7.6 &1.36 &2.96 &2.9 &2.25 &4.04 &5.26 \\
TCN &2.74 &5.68 &6.54 &1.45 &3.01 &3.03 &3.25 &5.51 &6.7 \\
DeepState &2.72  &5.24 &6.8 &1.88 &3.04 &2.8 &3.95 &6.49 &7.9 \\
GraphWaveNet &2.69 &5.15 &6.9 &1.3 &2.74 &2.7 &- &- &- \\
DeepGLO &2.91 &5.48 &6.75 &1.39 &2.91 &3.01 &3.01 &5.25 &6.2 \\
StemGNN &2.56 &5.06 &6.46 &1.23 &2.48 &2.63  &2.14  &4.01  &5.01 \\
AST &2.35 &4.84 &5.67 &1.25 &2.62 &2.53 &1.91 &3.60 &4.29 \\
Pyraformer &2.51  &4.88  &6.19 &1.24 &2.49 &2.57 &2.02 &3.68 &4.55  \\
FEDformer &2.87 &4.84 &6.70 &1.59 &2.76 &3.20 &2.52 &3.91 &5.57  \\
DLinear &2.67 &4.96 &6.54 &1.38 &2.84 &2.84 &2.00 &3.58 &4.57 \\
PatchTST &2.48 &4.89 &6.09 &1.27 &2.62 &2.61 &1.90 &3.52 &4.27 \\
Corrformer &2.48 &4.81  &6.03 &1.28 &2.58 &2.61 &1.92 &3.51 &4.34\\
MGCP & \textbf{2.12} & \textbf{3.82} & \textbf{4.98} & \textbf{1.00} & \textbf{1.92} & \textbf{2.00} & \textbf{1.61} & \textbf{2.83} & \textbf{3.58}\\
\hline
\hline 
Datasets&      &PEMS03     &  &      & PEMS04     & &      & PEMS08     &            
 \\
 \hline
Metrics &\text{  }MAE\text{  } &\text{  }RMSE\text{ }  &MAPE(\%) &\text{  }MAE\text{  } &\text{  }RMSE\text{ }  &MAPE(\%)&\text{  }MAE\text{  } &\text{  }RMSE\text{ }  &MAPE(\%)
     \\
\hline
ARIMA &16.77 &25.80&19.07 &22.08 &34.35 &14.87 &17.58 &27.15 &10.\\
FC-LSTM &21.33 &35.11 &23.33 &27.14 &41.59 &18.2 &22.2 &34.06 &14.2\\
SFM &17.67 &30.01 &18.33 &24.36 &37.10 &17.2 &16.01 &27.41 &10.4\\
N-BEATS &18.45 &31.23 &18.35 &25.56 &39.9 &17.18 &19.48 &28.32 &13.5\\
DCRNN &18.18 &30.31 &18.91 &24.7 &38.12 &17.12 &17.86 &27.83 &11.45\\
LSTNet &19.07 &29.67 &17.73 &24.04 &37.38 &17.01 &20.26 &31.96 &11.3\\
ST-GCN &17.49 &30.12 &17.15 &22.70 &35.50 &14.59 &18.02 &27.83 &11.4\\
TCN &18.23 &25.04 &19.44 &26.31 &36.11 &15.62 &15.93 &25.69 &16.5\\
DeepState &15.59 &20.21 &18.69 &26.5 &33.0 &15.4 &19.34 &27.18 &16\\
GraphWaveNet &19.85 &32.94 &19.31 &26.85 &39.7 &17.29 &19.13 &28.16 &12.68\\
DeepGLO &17.25 &23.25 &19.27 &25.45 &35.9 &12.2 &15.12 &25.22 &13.2\\
StemGNN &14.32 &21.64 &16.24 &20.24 &32.15 &\textbf{10.03} &15.83 &24.93 &9.26\\
AST &13.75 &21.43 &13.20 &19.21 &30.84 &12.45
&15.04 &23.55 &9.29\\
Pyraformer &13.76 &21.22 &14.28 &19.24  &30.54  &13.16  &15.43  &23.84  &9.97 \\
FEDformer &13.50  &20.85  &15.08 &19.03 & \textbf{29.98} &13.61 &14.60 &22.43 &9.44  \\
DLinear &15.37 &23.57 &16.19 &20.78 &32.34 &15.22 &16.34 &24.93 &10.30 \\
PatchTST  &14.96  &22.94  &14.35 &20.16 &31.71 &13.36 &15.50 &24.03 &9.71 \\
MGCP & \textbf{12.56} & \textbf{19.62} & \textbf{12.53} & \textbf{18.74} &30.10 &12.39 & \textbf{12.94} & \textbf{20.54} & \textbf{8.13}\\
\hline



\end{tabular}
\label{tab:experiments}
\end{footnotesize}
\end{table*}

Secondly, we notice that FC-LSTM and SFM perform almost the worst on most datasets. We believe this is because these two algorithms are based on very simple LSTM networks, which indicates it is difficult for LSTM to learn the complex interrelationships for different time steps of time series.

Thirdly, compared with LSTM-based methods, CNN-based models, N-BEATS and DeepState achieve obviously better performance for all the datasets. Since these methods adopt either convolution layer, or fully connected layer instead of only using LSTM layers, it indicates convolution and fully connected networks are more suitable to learn interrelationships for long-range time series.

In addition, it is observed that DCRNN, ST-GCN, and GraphWaveNet achieve much better performance than those mentioned in the previous three paragraphs, by using graph relationships between multiple time series. This demonstrates it can effectively reduce the learning complexity to incorporate the graph relation. However, these methods are not applicable for datasets without graph relation, which limits their generality.

Finally, the state-of-the-art AST \cite{wu2020adversarial}, Pyraformer\cite{liu2022pyraformer}, FEDformer \cite{zhou2022fedformer}, DLinear \cite{zeng2023transformers}, PatchTST \cite{DBLP:conf/iclr/NieNSK23} and StemGNN\cite{cao2021spectral} algorithms are significantly superior to the previous methods on almost all datasets, which demonstrates the excellent ability of attention-based models in learning temporal correlations, and also illustrates the importance of learning the inter-series correlations of multivariate time series through GNN. 

Our method is capable of learning the correlations of multivariate time series data at multi-grained levels, which covers the strengths of previous methods and results in improved performance. 
Specifically, MGCP surpasses the second-best baselines, reducing the MAE, RMSE and MAPE by $11.50\%$, $12.25\%$ and $7.17\%$ respectively in traffic datasets. Furthermore, unlike some methods that rely on graph structure information, such as DCRNN and ST-GCN, or those that depend on location information, like Corrformer, MGCP does not require these information, making it more universally applicable.

\begin{table*}
\centering
\begin{footnotesize}

\caption{Results of partial methods for three datasets.}
\renewcommand\arraystretch{1.35}
\tabcolsep=0.05cm
\begin{tabular}{lccc|ccc|ccc}
\hline
Datasets &      &Solar     &  &      &Electricity    & &      & ECG5000     &            \\
\hline
Metrics &MAE &RMSE   &MAPE(\%) &MAE &RMSE   &MAPE(\%)&MAE &RMSE   &MAPE(\%)
     \\
\hline
\text{StemGNN} & 0.0499 & 0.1267 & 15.02 & 0.0323 & 0.0515 & 11.13 & 0.0542 & 0.0806 & 19.78 \\
\text{AST} & 0.0510 & 0.1378 & 15.44 & 0.0282 & 0.0489 & 9.49 & 0.0535 & 0.0805 & 18.99 \\
\text{Pyraformer} & 0.0658 & 0.1355 & 39.03 & 0.0332 & 0.0535 & 11.62 & 0.0669 & 0.0917 & 22.90 \\
\text{FEDformer} & 0.0902 & 0.1568 & 53.37 & 0.0380 & 0.0541 & 14.64 & 0.0559 & 0.0826 & 19.51 \\
\text{DLinear} & 0.0855 & 0.1548 & 50.88 & 0.0380 & 0.0622 & 13.77 & 0.0538 & 0.0803 & 19.29 \\
\text{PatchTST} & 0.0536 & 0.1327 & 21.08 & 0.0301 & 0.0503 & 9.80 & 0.0531 & 0.0799 & 18.90 \\
\text{MGCP} & \textbf{0.0433} & \textbf{0.1038} & \textbf{14.68} & \textbf{0.0275} & \textbf{0.0441} & \textbf{9.09} & \textbf{0.0505} & \textbf{0.0759} & \textbf{18.75}\\

\hline
\end{tabular}
\label{tab:experiments2}
\end{footnotesize}
\end{table*}

\subsection{Ablation Study}
To understand the contribution of each component to MGCP, we conduct ablation experiments. First, we study the contribution of latent space learning to our model, including AFNO and GCN, and second we study the contribution of GANs to our model. Finally, we replace attention-based predictor(AP) with LSTM to test the contribution of this attention-based predictor. Results of all ablation experiments are shown in Table \ref{tab:ablation} and Table \ref{tab:ablation2}, from which we can make some observations.

Firstly, we can observe that MGCP without AFNO suffers from evidently larger MAE, RMSE, and MAPE for all the datasets. 
This is because AFNO can learn the global spatiotemporal correlations, which is beneficial to the extraction of multivariate time series features by subsequent components.

Secondly, without using GCN, the performance of MGCP drops significantly. 
This is because GCN can learn the inter-series correlations of the multivariate time series, which can greatly improve the prediction performance of MGCP.

\begin{table*}[t]
\centering
\begin{scriptsize}
\caption{Results of ablation experiments for traffic datasets}
\renewcommand\arraystretch{1.2}
\tabcolsep=0.07cm
\begin{tabular}{lccc|ccc|ccc}
\hline
Datasets &      &METR-LA     &  &      & PEMS-BAY     & &      & PEMS07     &            \\
\hline
Metrics &MAE &RMSE   &MAPE(\%) &MAE &RMSE   &MAPE(\%)&MAE &RMSE   &MAPE(\%)
     \\
\hline
MGCP & \textbf{2.12$\pm$0.03} & \textbf{3.82$\pm$0.04} & \textbf{4.98$\pm$0.08} & \textbf{1.00$\pm$0.01} & \textbf{1.92$\pm$0.01} & \textbf{2.00$\pm$0.02} & \textbf{1.61$\pm$0.04} & \textbf{2.83$\pm$0.11} & \textbf{3.58$\pm$0.13}\\
w/o GANs &2.14$\pm$0.08 & 3.83$\pm$0.14 & 5.01$\pm$0.22 &1.01$\pm$0.01 & 1.93$\pm$0.03 & 2.02$\pm$0.02 &1.63$\pm$0.05 & 2.87$\pm$0.09 & 3.63$\pm$0.11\\
w/o GCN &2.49$\pm$0.01 & 4.74$\pm$0.01 & 6.14$\pm$0.02 &1.23$\pm$0.00 & 2.51$\pm$0.00 & 2.55$\pm$0.01 &1.87$\pm$0.01 & 3.40$\pm$0.00 & 4.24$\pm$0.01\\
w/o AFNO &2.40$\pm$0.01 & 4.45$\pm$0.02 & 5.82$\pm$0.03 &1.19$\pm$0.00 & 2.30$\pm$0.03 & 2.42$\pm$0.01 &1.82$\pm$0.01 & 3.22$\pm$0.02 & 4.08$\pm$0.03\\
w/o AP &2.17$\pm$0.06 & 3.93$\pm$0.08 & 5.16$\pm$0.15 &1.07$\pm$0.04 & 2.06$\pm$0.08 & 2.15$\pm$0.08 &1.69$\pm$0.08 & 2.98$\pm$0.14 & 3.78$\pm$0.17 \\
\hline
\hline
Datasets &      &PEMS03     &  &      & PEMS04     & &      & PEMS08     &            \\
\hline
Metrics &MAE &RMSE   &MAPE(\%) &MAE &RMSE   &MAPE(\%)&MAE &RMSE   &MAPE(\%)
     \\
\hline
MGCP &\textbf{12.56$\pm$0.20} & \textbf{19.62$\pm$0.39} & \textbf{12.53$\pm$0.38} & \textbf{18.74$\pm$0.30} & \textbf{30.10$\pm$0.51} & \textbf{12.39$\pm$0.23} & \textbf{12.94$\pm$0.12} & \textbf{20.54$\pm$0.16} & \textbf{8.13$\pm$0.06}\\
w/o GANs &12.79$\pm$0.56 & 19.94$\pm$0.90 & 12.67$\pm$0.73&18.99$\pm$0.11 & 30.51$\pm$0.19 & 12.60$\pm$0.07 &12.98$\pm$0.10 & 20.59$\pm$0.15 & 8.15$\pm$0.07 \\
w/o GCN &13.59$\pm$0.01 & 21.20$\pm$0.04 & 13.67$\pm$0.02 &19.04$\pm$0.01 & 30.56$\pm$0.02 & 12.63$\pm$0.01 &14.12$\pm$0.02 & 22.46$\pm$0.04 & 8.92$\pm$0.01\\
w/o AFNO &13.63$\pm$0.03 & 21.19$\pm$0.08 & 13.61$\pm$0.14 &18.94$\pm$0.04 & 30.37$\pm$0.06 & 12.60$\pm$0.08 &14.04$\pm$0.07 & 22.18$\pm$0.07 & 8.84$\pm$0.04\\
w/o AP &12.90$\pm$0.20 & 20.08$\pm$0.33 & 12.71$\pm$0.31 &19.21$\pm$0.04 & 30.90$\pm$0.08 & 13.05$\pm$0.10 &13.43$\pm$0.07 & 21.21$\pm$0.09 & 8.41$\pm$0.03\\
\hline
\end{tabular}
\label{tab:ablation}
\end{scriptsize}
\end{table*}
\begin{table*}[t]
\begin{tiny}
\centering
\caption{Results of ablation experiments for three datasets.}
\renewcommand\arraystretch{1.35}
\tabcolsep=0.05cm
\begin{tabular}{lccc|ccc|ccc}
\hline
Datasets &      &Solar     &  &      &Electricity    & &      & ECG5000     &            \\
\hline
Metrics &MAE &RMSE   &MAPE(\%) &MAE &RMSE   &MAPE(\%)&MAE &RMSE   &MAPE(\%)
     \\
\hline
MGCP &\textbf{0.0433$\pm$0.0008} &\textbf{0.1038$\pm$0.0017} &\textbf{14.68$\pm$0.93} &\textbf{0.0275$\pm$0.0017} & \textbf{0.0441$\pm$0.0026} & \textbf{9.09$\pm$0.67} &\textbf{0.0505$\pm$0.0005} & \textbf{0.0759$\pm$0.0011} & \textbf{18.75$\pm$0.27}\\
w/o GANs &0.0435$\pm$0.0006 &0.1044$\pm$0.0008 &15.68$\pm$0.91 &0.0279$\pm$0.0014 & 0.0446$\pm$0.0024 & 9.28$\pm$0.67 &0.0530$\pm$0.0007 & 0.0801$\pm$0.0010 & 19.41$\pm$0.40\\
w/o GCN&0.0474$\pm$0.0004 &0.1182$\pm$0.0002 &16.28$\pm$0.59 &0.0284$\pm$0.0001 & 0.0455$\pm$0.0001 & 8.89$\pm$0.02 &0.0524$\pm$0.0001 & 0.0794$\pm$0.0001 & 19.00$\pm$0.06\\
w/o AFNO & 0.0552$\pm$0.0003 &0.1218$\pm$0.0005 &24.71$\pm$1.44 &0.0298$\pm$0.0001 & 0.0486$\pm$0.0002 & 9.90$\pm$0.33 &0.0519$\pm$0.0006 & 0.0785$\pm$0.0010 & 18.82$\pm$0.29\\
w/o AP &0.0455$\pm$0.0010 &0.1085$\pm$0.0026 &15.74$\pm$1.51 &0.0300$\pm$0.0013 & 0.0473$\pm$0.0022 & 9.98$\pm$0.18 &0.0514$\pm$0.0007 & 0.0771$\pm$0.0009 & 19.07$\pm$0.38\\
\hline
\end{tabular}
\label{tab:ablation2}
\end{tiny}
\end{table*}
Thirdly, we observed that the removal of GANs in MGCP resulted in a slight decrease in performance for some datasets, but a significant decrease for others, such as PEMS03, PEMS04, and ECG5000. This outcome is unsurprising, as in supervised tasks, the optimization objective shown in Equation \eqref{Eq:goal} can already produce results that are very close to the actual values in most cases. The optimization objective shown in Equation \eqref{Eq:goal2} generally only serves as an auxiliary role in supervised tasks. Therefore, GANs have played a crucial role in achieving a breakthrough, namely, making the predicted data distribution more closely approximate the distribution of actual data.

Finally, the LSTM based MGCP performs significantly worse than the attention based MGCP model, which indicates the attention mechanism is much better for learning on long-range time series.
By the way, it can be seen that the use of the predictor component of MGCP is very flexible, so that MGCP still has room for improvement in the future.


\begin{figure*}[htbp]
\centering  
\includegraphics[width=0.49\textwidth]{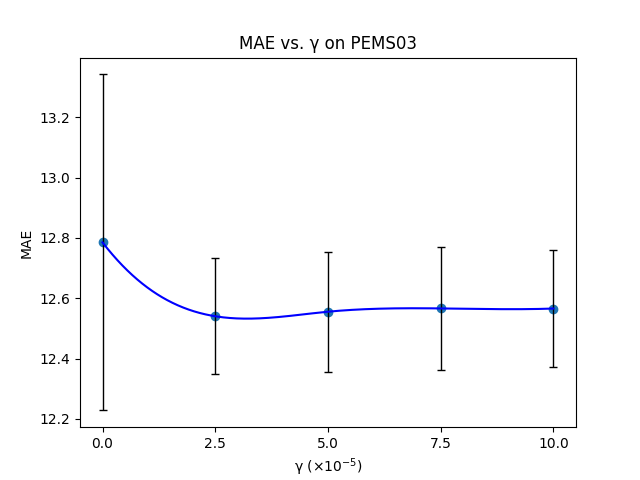}
\includegraphics[width=0.49\textwidth]{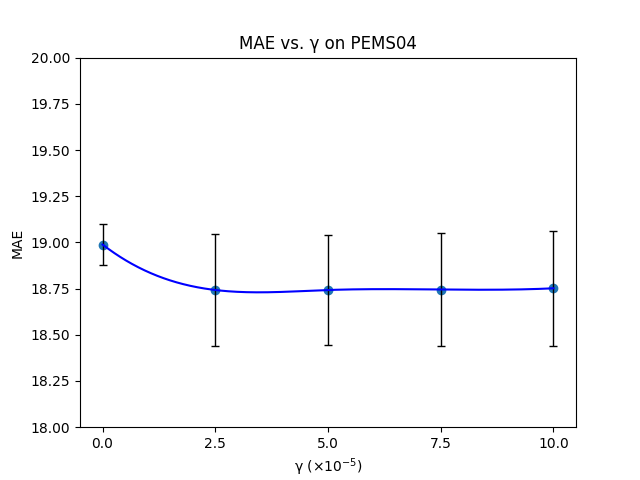}
\caption{
Evaluation of the performance with respect to varied $\gamma$ over partial datasets. The blue dots represent the average test error over multiple experiments, and the black lines represent the corresponding variances.
}\label{fig:gamma}
\end{figure*}
\subsection{Parametric Analysis}
Fig. \ref{fig:gamma} evaluates the performance of MGCP with varied hyperparameter $\gamma$ over partial datasets.
For this experiment, we let $\gamma=[0,2.5,5,7.5,10]\times10^{-5}$, and fix the other hyperparameters to be consistent with the experiment in Table \ref{tab:experiments}, e.g. $\beta=0.9$ and $\lambda=0.1$. Firstly, when the value of $\gamma$ is zero, the average test error is large, which indicates that adversarial learning can improve the performance of MGCP. Secondly, When $\gamma$ fluctuates 50\% up and down on a benchmark of $5\times10^{-5}$, the performance of MGCP fluctuates by $0.12\%$ in the PEMS03 dataset and $0.09\%$ in the PEMS04 dataset, which indicates the MGCP model is very stable with respect to the hyperparameter $\gamma$.

\begin{figure*}[htbp]
\centering  
\includegraphics[width=0.49\textwidth]{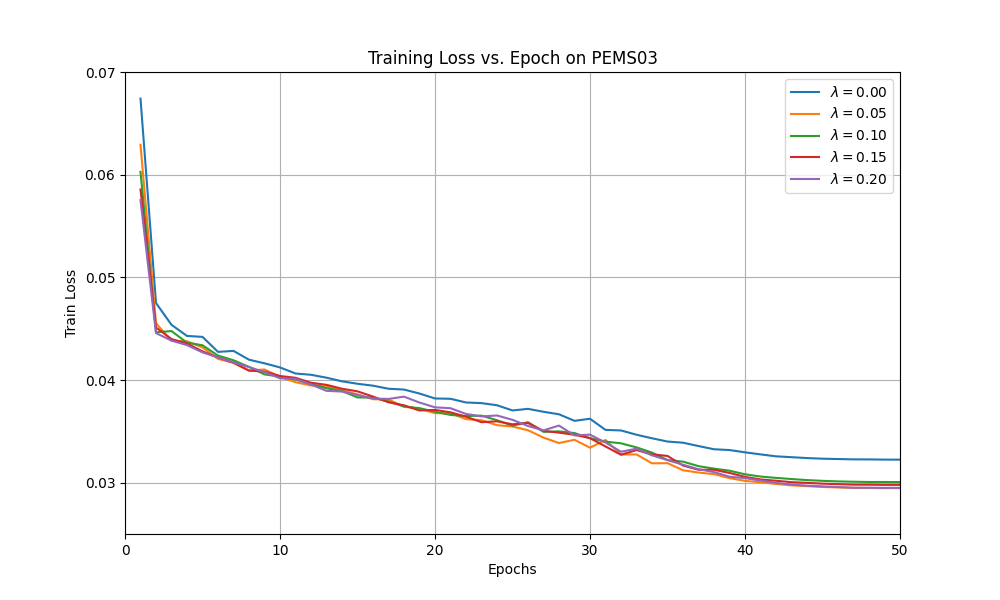}
\includegraphics[width=0.49\textwidth]{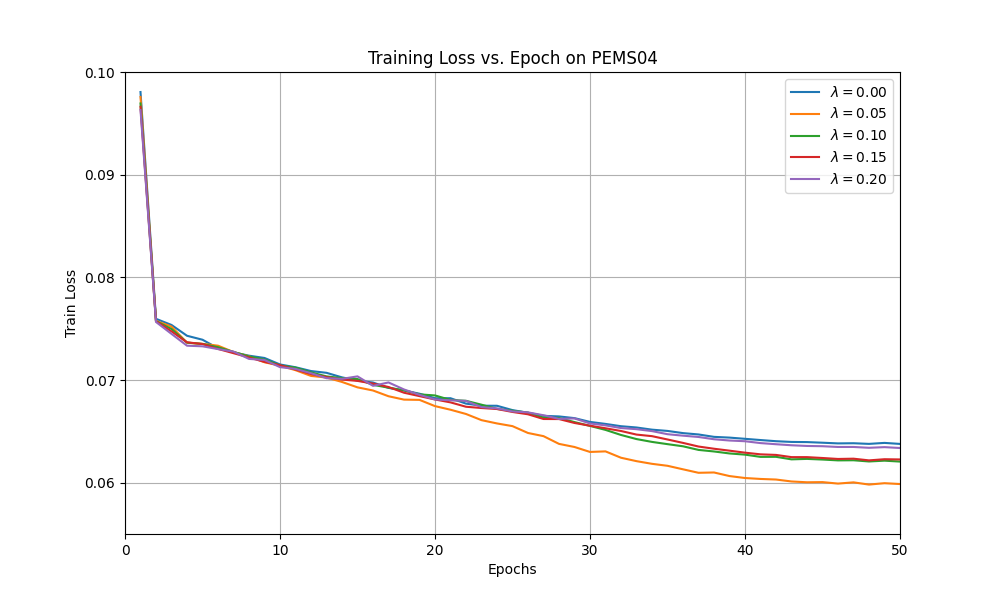}
\caption{How training loss vary with epochs under different $\lambda$.}
\label{fig:lambda}
\end{figure*}

Fig. \ref{fig:lambda} evaluates the performance of MGCP with varied hyperparameter $\lambda$ over partial datasets.
For this experiment, we let $\lambda=[0,0.05,0.1,0.15,0.2]$, and fix the other hyperparameters to be consistent with the experiment in Table \ref{tab:experiments}, e.g., $\gamma=5\times 10^{-5}, \beta=0.9$.
Firstly, we observe that the convergence value of the train loss is worst when $\lambda=0$, which indicates the importance of the reconstruction loss. Secondly, $\lambda$ has a stable interval for each data set, within which the convergence value are generally better. For example, under the PEMS03 dataset, the interval is [0.05,0.2], and under the PEMS04 dataset  it is [0.05,0.15].

\begin{figure*}[htbp]
\centering  
\includegraphics[width=0.49\textwidth]{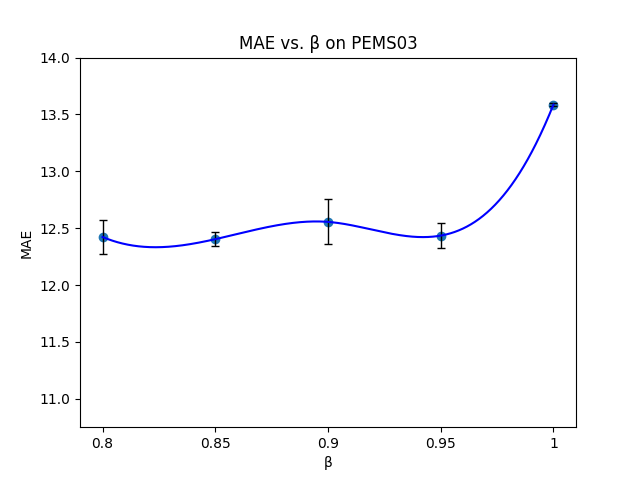}
\includegraphics[width=0.49\textwidth]{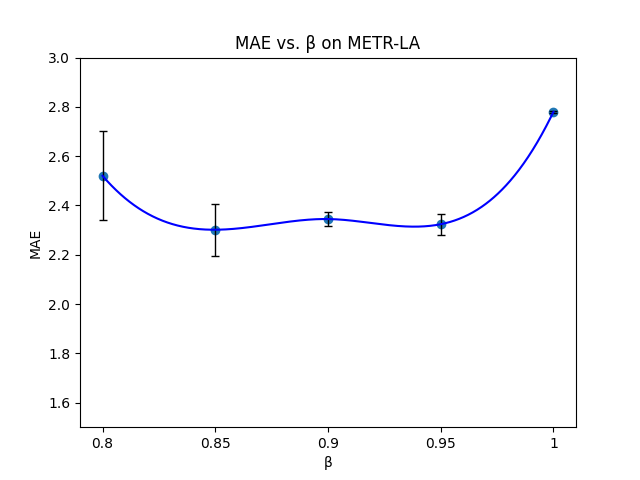}
\caption{
Evaluation of the performance with respect to varied $\beta$ over partial dataset. Dots and lines have the same meaning as in Fig. \ref{fig:gamma}.}
\label{fig:beta}
\end{figure*}

Fig. \ref{fig:beta} evaluates the performance of MGCP with varied hyperparameter $\beta$ over partial datasets.
For this experiment, let $\beta=[0.8,0.85,0.9,0.95,1]$, $\lambda=0.1$, $\gamma=5\times10^{-5}$ for the PEMS03 dataset, and $\gamma=2\times10^{-6}$ for METR-LA dataset. Firstly, when $\beta=1$, i.e., the adjacency matrix is an identity matrix, the test loss is significantly larger than when $\beta=[0.85,0.9,0.95]$, which indicates the GCN can indeed help MGCP learn the inter-series correlations of multivariate time series. Secondly, when the $\beta$ is small enough, the performance may deteriorate. For example, when $\beta=0.8$, the average test error and the variance is large.  We believe that this is because too few weights of the identity matrix in the adjacency matrix will cause the time series after multi-layer graph convolutions to lose their own features too much, i.e., oversmoothing problem in GNN.

\begin{table*}[t]
\begin{scriptsize}
\centering
\caption{ Performance comparison for longer prediction steps.}
\renewcommand\arraystretch{1.2}
\tabcolsep=0.07cm
\begin{tabular}{lccc|ccc|ccc}

\hline
PEMS03 &       &15min     &  &      &30min  & &      & 1hour    & \\
\hline
Metrics &MAE &RMSE   &MAPE(\%) &MAE &RMSE   &MAPE(\%)&MAE &RMSE   &MAPE(\%)
     \\
\hline
StemGNN &14.32$\pm$0.01 &21.64$\pm$0.16 &16.24$\pm$0.06 &15.73$\pm$0.20	&25.71$\pm$0.08 &16.39$\pm$0.80 &19.87$\pm$	3.04 &31.46$\pm$3.64	&22.71$\pm$5.79\\
AST &13.75$\pm$0.04 &21.43$\pm$0.06 &13.20$\pm$0.13 & 14.50$\pm$0.03 &23.06$\pm$0.07 &14.35$\pm$0.05
&16.96$\pm$0.05 &27.20$\pm$0.22 &\textbf{16.33$\pm$0.04}\\
Pyraformer &13.76$\pm$0.02  &21.22$\pm$0.03  &14.28$\pm$0.06 &14.97$\pm$0.01 &23.37$\pm$0.01 &15.31$\pm$0.17 &17.18$\pm$0.03 &27.05$\pm$0.03    &17.74$\pm$0.07  \\
FEDformer &13.50$\pm$0.02  &20.85$\pm$0.05  &15.08$\pm$0.14 &14.43$\pm$0.01 &22.48$\pm$0.02 &15.84$\pm$0.06 &17.20$\pm$0.02  &\textbf{26.23$\pm$0.01}  &18.06$\pm$0.19 \\
DLinear &15.37$\pm$0.03 &23.57$\pm$0.01 &16.19$\pm$0.23 &17.55$\pm$0.05 &27.24$\pm$0.01 &19.51$\pm$0.32 &21.80$\pm$0.04 &34.10$\pm$0.01 &25.46$\pm$0.27 \\
PatchTST &14.96$\pm$0.01  &22.94$\pm$0.01  &14.35$\pm$0.02 &16.99$\pm$0.02 &26.50$\pm$0.03 &15.98$\pm$0.04 &21.03$\pm$0.04  &33.44$\pm$0.03  &19.30$\pm$0.06 \\
ATP &\textbf{12.56$\pm$0.20} & \textbf{19.62$\pm$0.39} & \textbf{12.53$\pm$0.38} & \textbf{13.76$\pm$0.22} & \textbf{21.79$\pm$0.44} & \textbf{13.51$\pm$0.38} & \textbf{16.82$\pm$0.11} & 27.23$\pm$0.38 & 16.42$\pm$0.11\\

\hline
PEMS07 &       &15min     &  &      &30min  & &      & 1hour    & \\
\hline
Metrics &MAE &RMSE   &MAPE(\%) &MAE &RMSE   &MAPE(\%)&MAE &RMSE   &MAPE(\%)
     \\
\hline
StemGNN  &2.14$\pm$0.17  &4.01$\pm$0.18  &5.01$\pm$0.36 &2.86$\pm$0.38	&4.86$\pm$0.32	&6.73$\pm$0.72 &3.62$\pm$0.17	&6.29$\pm$0.21	&9.21$\pm$0.47
 \\
AST &1.91$\pm$0.01 &3.60$\pm$0.01 &4.29$\pm$0.02 &2.43$\pm$0.01 &4.76$\pm$0.02 &5.61$\pm$0.02 &3.31$\pm$0.04 &6.59$\pm$0.08 &7.80$\pm$0.09\\
Pyraformer &2.02$\pm$0.01 &3.68$\pm$0.01 &4.55$\pm$0.02 &2.56$\pm$0.01  &4.88$\pm$0.01  &6.01$\pm$0.01 &3.43$\pm$0.01  &6.47$\pm$0.06  &8.45$\pm$0.05\\
FEDformer &2.52$\pm$0.01 &3.91$\pm$0.01 &5.57$\pm$0.02 &3.22$\pm$0.01 &5.10$\pm$0.01 &7.23$\pm$0.02 &4.09$\pm$0.03 &6.45$\pm$0.03 &9.31$\pm$0.10 \\
DLinear &2.00$\pm$0.01 &3.58$\pm$0.01 &4.57$\pm$0.01 &2.66$\pm$0.02 &4.88$\pm$0.01 &6.33$\pm$0.01 &3.72$\pm$0.02	&6.61$\pm$0.01	&9.23$\pm$0.01\\
PatchTST &1.90$\pm$0.01 &3.52$\pm$0.01 &4.27$\pm$0.01 &2.51$\pm$0.01 &4.86$\pm$0.01 &5.80$\pm$0.02 &3.44$\pm$0.01 &6.74$\pm$0.01 &8.28$\pm$0.01\\
Corrformer &1.92$\pm$0.01 &3.51$\pm$0.01 &4.34$\pm$0.01 &2.52$\pm$0.01 &4.78$\pm$0.01 &5.82$\pm$0.01 &3.37$\pm$0.01 &6.41$\pm$0.01 &7.98$\pm$0.02\\
MGCP &\textbf{1.61$\pm$0.04} & \textbf{2.83$\pm$0.11} & \textbf{3.58$\pm$0.13} & \textbf{2.21$\pm$0.18} & \textbf{4.11$\pm$0.38} & \textbf{5.11$\pm$0.49} & \textbf{3.26$\pm$0.12} & \textbf{6.09$\pm$0.25} & \textbf{7.74$\pm$0.31}\\

\hline
PEMS08 &       &15min     &  &      &30min  & &      & 1hour    & \\
\hline
Metrics &MAE &RMSE   &MAPE(\%) &MAE &RMSE   &MAPE(\%)&MAE &RMSE   &MAPE(\%)
     \\
\hline
StemGNN &15.83$\pm$0.18 &24.93$\pm$0.18 &9.26$\pm$0.17 &17.05$\pm$0.27 &26.06$\pm$0.36	&11.09$\pm$0.31 &20.02$\pm$0.51	&30.37$\pm$0.81	&13.10$\pm$0.21
\\
AST  &15.04$\pm$0.01 &23.55$\pm$0.01 &9.29$\pm$0.02 &15.24$\pm$0.03 &24.40$\pm$0.06 &9.39$\pm$0.02 &18.27$\pm$0.52	&29.20$\pm$0.87	&11.04$\pm$0.28
\\
Pyraformer  &15.43$\pm$0.12  &23.84$\pm$0.12  &9.97$\pm$0.11 &16.60$\pm$0.06  &26.19$\pm$0.06 &10.74$\pm$0.06 &19.41$\pm$0.04  &30.76$\pm$0.04  &12.73$\pm$0.11\\
FEDformer &14.60$\pm$0.03  &22.43$\pm$0.01  &9.44$\pm$0.08 &15.58$\pm$0.03 &24.08$\pm$0.04 &10.11$\pm$0.05 &17.72$\pm$0.08  &\textbf{27.36$\pm$0.06}  &11.47$\pm$0.10 \\
DLinear &16.34$\pm$0.01  &24.93$\pm$0.01  &10.30$\pm$0.03 &18.54$\pm$0.05  &28.42$\pm$0.01  &11.88$\pm$0.17 &22.98$\pm$0.08  &34.91$\pm$0.03  &15.42$\pm$0.20 \\
PatchTST &15.59$\pm$0.01 &24.03$\pm$0.01 &9.71$\pm$0.01 &17.45$\pm$0.01 &27.39$\pm$0.01 &10.84$\pm$0.01 &21.45$\pm$0.02	&33.86$\pm$0.01	&13.23$\pm$0.03\\
MGCP &\textbf{12.94$\pm$0.12} & \textbf{20.54$\pm$0.16} & \textbf{8.13$\pm$0.06} & \textbf{14.22$\pm$0.18} & \textbf{22.95$\pm$0.27} & \textbf{8.87$\pm$0.09} & \textbf{17.26$\pm$0.30} & 28.12$\pm$0.49 & \textbf{10.68$\pm$0.21}\\
\hline
\end{tabular}
\label{tab:long step}
\end{scriptsize}
\end{table*}

\subsection{Long Predicted Steps Analysis}
In this part, we further investigate the performance of different methods in multivariate time series forecasting with longer predicted steps. Specifically, We increase the predicted steps from the original 15min to 30min and 60min to challenge the long-range predictive ability of the models, and Table \ref{tab:long step} shows the performance of some methods at different predicted steps under the datasets PEMS07, PEMS08 and PEMS03.

Table \ref{tab:long step} reveals several key observations. First, in the prediction of longer time series, StemGNN underperforms compared to some attention-based models, which indirectly highlights the superiority of attention mechanisms in modeling long-range temporal correlations. Additionally, our experimental results demonstrate that our proposed method outperforms the other methods in most cases, indicating its effectiveness in handling long-term time series data. However, in a few instances, such as the PEMS03 dataset with a predicted step of 1 hour, MGCP is outperformed by certain indicators. We believe that over longer time series, the temporal correlations become more significant than the inter-series correlations and the global spatiotemporal correlations. Therefore, our predictor with only one layer of attention-based encoder-decoder structure may be weaker than AST with three layers of attention-based encoder-decoder structure in learning long-term temporal correlations, and FEDformer can better handle long time series in some cases because of its decomposition scheme. 

\subsection{Mixers Analysis}
\begin{table}[t]
\centering
\caption{Performance and Cost Comparison of different mixers}
\renewcommand\arraystretch{1.25}
\tabcolsep=0.12cm
\begin{tabular}{c|ccc}
\hline
\multicolumn{1}{c|}{Mixer}&\multicolumn{1}{c}{MAPE} & \multicolumn{1}{c}{Time/} & \multicolumn{1}{c}{Memory}\\
Name &(\%) &Epoch (s) &Usage (MB) \\
\hline
Global Self-Attention   &8.90$\pm$0.15 &1060 &7026       \\
FNO &9.70$\pm$0.20 &1224 &5544   \\
AFNO  &\textbf{8.16$\pm$0.13} &\textbf{774} &\textbf{1616}        \\
\hline
\end{tabular}
\label{tab:mixers}
\end{table}

\begin{figure}[htbp]
\begin{center}
\includegraphics[width=0.95\textwidth]{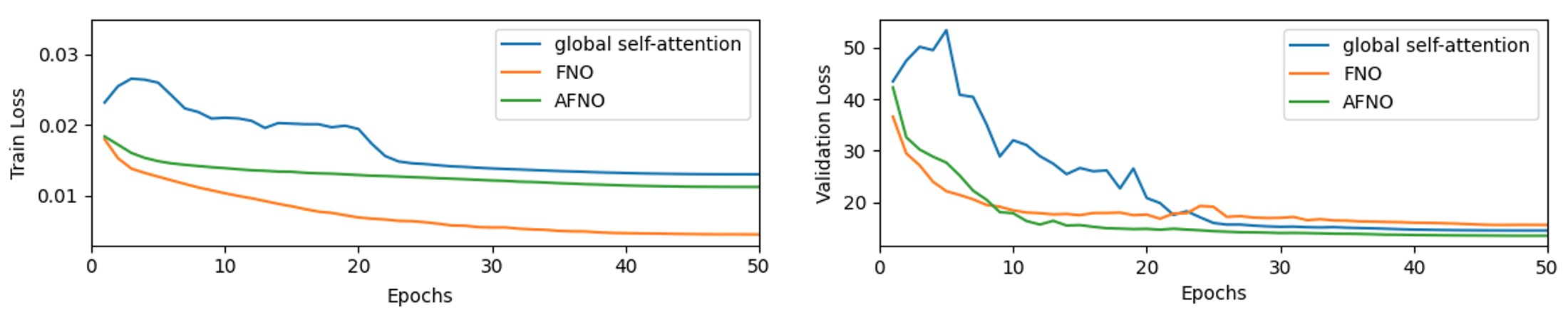}
\end{center}
\caption{How training loss and validation loss vary with epochs under different mixers.}

\label{fig:mixers}
\end{figure}
To analyze the advantages of AFNO, we replaced the AFNO in the MGCP with different mixers and then compared their performance and cost. Specifically, we compare global self-attention, FNO and AFNO in predicting performance, time spent per epoch, and the amount of memory used by the model. Because the spatial complexity of global self-attention is large, OOM will be caused under the default batch size. Therefore, for the sake of fairness, all mixers in this experiment perform and spend on the same and smaller batch size, resulting in slight discrepancies with Table \ref{tab:experiments}. In addition, when a mask is necessary, the FNO applies adjustments similar to those of Masked AFNO. Finally, except for the mixer, the other settings are exactly the same.

Table \ref{tab:mixers} shows the performance and cost of the model under different mixers, and Fig. \ref{fig:mixers} depicts the change curves of training loss and validation loss with epochs under different mixers.
From the results, AFNO mixers are optimal in terms of performance, time spent and space usage. In addition, we can also find that the training loss of the FNO can converge to a smaller amount, but its verification loss is the worst. We believe this is because FNO uses different channel mix weight parameters for all timestamps, resulting in a large number of parameters and easy overfitting. In general, it is very appropriate to use AFNO as a mixer in MGCP for learning the global spatiotemporal correlations.

\section{Conclusion}
In this paper, we introduce the MGCP network, a new model for multivariate time series prediction, which can learn the correlations at multi-grained levels to make accurate prediction. Specifically, at fine-grained level, MGCP utilizes attention mechanism and Adaptive Fourier Neural Operators to extract the long-term temporal correlations and the global spatiotemporal correlations, respective. At medium-grained level, MGCP uses Graph Convolution Network to learn the inter-series correlations. Furthermore, we also modify the discriminator into a conditional discriminator for better adaptation to time series prediction tasks. Through the adversarial training of the predictor and the conditional discriminator, MGCP can make the prediction results closer to real data from a distribution-level, i.e., learning the correlations at coarse-grained level.
Finally, we conduct experiments on a large amount of real datasets, demonstrating the efficiency and generality of MGCP. The ablation experiment and the parameter analysis experiment prove the rationality of design of MGCP components and the stability of MGCP model, respectively. 
Moreover, we analyze the performance of MGCP over longer forecast steps, and compare the effects of different mixers on both performance and cost. The results of these analyses provide further validation of the benefits of MGCP.
Due to the flexibility in the choice of the sequence predictors, MGCP still has great potential for improvement in the future to cope with increasingly complex data. For example, we can optimize our sequence predictor by referring to FEDformer's decomposition scheme to enhance its ability to predict longer sequences. We will explore further optimizations in the future.

\bibliographystyle{elsarticle-num}
\bibliography{references}

\begin{thebibliography}{10}
\expandafter\ifx\csname url\endcsname\relax
  \def\url#1{\texttt{#1}}\fi
\expandafter\ifx\csname urlprefix\endcsname\relax\def\urlprefix{URL }\fi
\expandafter\ifx\csname href\endcsname\relax
  \def\href#1#2{#2} \def\path#1{#1}\fi

\bibitem{gao2022earthformer}
Z.~Gao, X.~Shi, H.~Wang, Y.~Zhu, Y.~B. Wang, M.~Li, D.-Y. Yeung, Earthformer: Exploring space-time transformers for earth system forecasting, Advances in Neural Information Processing Systems 35 (2022) 25390--25403.

\bibitem{ge2023advancing}
C.~Ge, X.~Ding, Z.~Tong, L.~Yuan, J.~Wang, Y.~Song, P.~Luo, Advancing vision transformers with group-mix attention, arXiv preprint arXiv:2311.15157 (2023).

\bibitem{liu2016online}
C.~Liu, S.~C. Hoi, P.~Zhao, J.~Sun, Online arima algorithms for time series prediction, in: Thirtieth AAAI conference on artificial intelligence, 2016.

\bibitem{hyndman2008forecasting}
R.~Hyndman, A.~B. Koehler, J.~K. Ord, R.~D. Snyder, Forecasting with exponential smoothing: the state space approach, Springer Science \& Business Media, 2008.

\bibitem{BOLLERSLEV1986307}
T.~Bollerslev, Generalized autoregressive conditional heteroskedasticity, Journal of Econometrics 31~(3) (1986) 307--327.

\bibitem{sutskever2014sequence}
I.~Sutskever, O.~Vinyals, Q.~V. Le, Sequence to sequence learning with neural networks, in: Advances in neural information processing systems, 2014, pp. 3104--3112.

\bibitem{zhang2017stock}
L.~Zhang, C.~Aggarwal, G.-J. Qi, Stock price prediction via discovering multi-frequency trading patterns, in: Proceedings of the 23rd ACM SIGKDD international conference on knowledge discovery and data mining, 2017, pp. 2141--2149.

\bibitem{makridakis2018statistical}
S.~Makridakis, E.~Spiliotis, V.~Assimakopoulos, Statistical and machine learning forecasting methods: Concerns and ways forward, PloS one 13~(3) (2018) e0194889.

\bibitem{makridakis2018m4}
M.~Spyros, S.~Evangelos, A.~Vassilios, The m4 competition: Results, findings, conclusion and way forward, International Journal of Forecasting 34~(4) (2018) 802--808.

\bibitem{Oreshkin2020N-BEATS:}
B.~N. Oreshkin, D.~Carpov, N.~Chapados, Y.~Bengio, N-beats: Neural basis expansion analysis for interpretable time series forecasting, in: International Conference on Learning Representations, 2020.

\bibitem{bai2018empirical}
S.~Bai, J.~Z. Kolter, V.~Koltun, An empirical evaluation of generic convolutional and recurrent networks for sequence modeling, arXiv preprint arXiv:1803.01271 (2018).

\bibitem{rangapuram2018deep}
S.~S. Rangapuram, M.~W. Seeger, J.~Gasthaus, L.~Stella, Y.~Wang, T.~Januschowski, Deep state space models for time series forecasting, Advances in neural information processing systems 31 (2018) 7785--7794.

\bibitem{wu2020adversarial}
S.~Wu, X.~Xiao, Q.~Ding, P.~Zhao, W.~Ying, J.~Huang, Adversarial sparse transformer for time series forecasting (2020).

\bibitem{liu2022pyraformer}
S.~Liu, H.~Yu, C.~Liao, J.~Li, W.~Lin, A.~X. Liu, S.~Dustdar, Pyraformer: Low-complexity pyramidal attention for long-range time series modeling and forecasting, in: International Conference on Learning Representations, 2022.

\bibitem{zhou2022fedformer}
T.~Zhou, Z.~Ma, Q.~Wen, X.~Wang, L.~Sun, R.~Jin, Fedformer: Frequency enhanced decomposed transformer for long-term series forecasting, in: International Conference on Machine Learning, PMLR, 2022, pp. 27268--27286.

\bibitem{zeng2023transformers}
A.~Zeng, M.~Chen, L.~Zhang, Q.~Xu, Are transformers effective for time series forecasting?, in: Proceedings of the AAAI conference on artificial intelligence, Vol.~37, 2023, pp. 11121--11128.

\bibitem{DBLP:conf/iclr/NieNSK23}
Y.~Nie, N.~H. Nguyen, P.~Sinthong, J.~Kalagnanam, A time series is worth 64 words: Long-term forecasting with transformers, in: The Eleventh International Conference on Learning Representations, {ICLR} 2023, Kigali, Rwanda, May 1-5, 2023, 2023.

\bibitem{li2017diffusion}
Y.~Li, R.~Yu, C.~Shahabi, Y.~Liu, Diffusion convolutional recurrent neural network: Data-driven traffic forecasting, in: 6th International Conference on Learning Representations, {ICLR} 2018, Vancouver, BC, Canada, April 30 - May 3, 2018, Conference Track Proceedings, OpenReview.net, 2018.

\bibitem{yu2017spatio}
B.~Yu, H.~Yin, Z.~Zhu, Spatio-temporal graph convolutional networks: {A} deep learning framework for traffic forecasting, in: J.~Lang (Ed.), Proceedings of the Twenty-Seventh International Joint Conference on Artificial Intelligence, {IJCAI} 2018, July 13-19, 2018, Stockholm, Sweden, ijcai.org, 2018, pp. 3634--3640.

\bibitem{wu2019graph}
Z.~Wu, S.~Pan, G.~Long, J.~Jiang, C.~Zhang, Graph wavenet for deep spatial-temporal graph modeling, in: S.~Kraus (Ed.), Proceedings of the Twenty-Eighth International Joint Conference on Artificial Intelligence, {IJCAI} 2019, Macao, China, August 10-16, 2019, ijcai.org, 2019, pp. 1907--1913.

\bibitem{cao2021spectral}
D.~Cao, Y.~Wang, J.~Duan, C.~Zhang, X.~Zhu, C.~Huang, Y.~Tong, B.~Xu, J.~Bai, J.~Tong, et~al., Spectral temporal graph neural network for multivariate time-series forecasting, Advances in neural information processing systems 33 (2020) 17766--17778.

\bibitem{NEURIPS2021_cba0a4ee}
I.~O. Tolstikhin, N.~Houlsby, A.~Kolesnikov, L.~Beyer, X.~Zhai, T.~Unterthiner, J.~Yung, A.~Steiner, D.~Keysers, J.~Uszkoreit, M.~Lucic, A.~Dosovitskiy, Mlp-mixer: An all-mlp architecture for vision, Vol.~34, 2021, pp. 24261--24272.

\bibitem{NEURIPS2021_4cc05b35}
H.~Liu, Z.~Dai, D.~So, Q.~V. Le, Pay attention to mlps, Vol.~34, 2021, pp. 9204--9215.

\bibitem{lee-thorp-etal-2022-fnet}
J.~Lee-Thorp, J.~Ainslie, I.~Eckstein, S.~Ontanon, {FN}et: Mixing tokens with {F}ourier transforms, in: Proceedings of the 2022 Conference of the North American Chapter of the Association for Computational Linguistics: Human Language Technologies, Association for Computational Linguistics, Seattle, United States, 2022, pp. 4296--4313.

\bibitem{NEURIPS2021_07e87c2f}
Y.~Rao, W.~Zhao, Z.~Zhu, J.~Lu, J.~Zhou, Global filter networks for image classification, in: M.~Ranzato, A.~Beygelzimer, Y.~Dauphin, P.~Liang, J.~W. Vaughan (Eds.), Advances in Neural Information Processing Systems, Vol.~34, Curran Associates, Inc., 2021, pp. 980--993.

\bibitem{guibas2022efficient}
J.~Guibas, M.~Mardani, Z.~Li, A.~Tao, A.~Anandkumar, B.~Catanzaro, Efficient token mixing for transformers via adaptive fourier neural operators, in: International Conference on Learning Representations, 2022.

\bibitem{goodfellow2014generative}
I.~Goodfellow, J.~Pouget-Abadie, M.~Mirza, B.~Xu, D.~Warde-Farley, S.~Ozair, A.~Courville, Y.~Bengio, Generative adversarial nets, Advances in neural information processing systems 27 (2014).

\bibitem{mogren2016c}
O.~Mogren, C-rnn-gan: Continuous recurrent neural networks with adversarial training, arXiv preprint arXiv:1611.09904 (2016).

\bibitem{yoon2019time}
J.~Yoon, D.~Jarrett, M.~Van~der Schaar, Time-series generative adversarial networks (2019).

\bibitem{lin2020using}
Z.~Lin, A.~Jain, C.~Wang, G.~Fanti, V.~Sekar, Using gans for sharing networked time series data: Challenges, initial promise, and open questions, in: Proceedings of the ACM Internet Measurement Conference, 2020, pp. 464--483.

\bibitem{zhang2023stad}
Z.~Zhang, W.~Li, W.~Ding, L.~Zhang, Q.~Lu, P.~Hu, T.~Gui, S.~Lu, Stad-gan: unsupervised anomaly detection on multivariate time series with self-training generative adversarial networks, ACM Transactions on Knowledge Discovery from Data 17~(5) (2023) 1--18.

\bibitem{lai2018modeling}
G.~Lai, W.-C. Chang, Y.~Yang, H.~Liu, Modeling long-and short-term temporal patterns with deep neural networks, in: The 41st International ACM SIGIR Conference on Research \& Development in Information Retrieval, 2018, pp. 95--104.

\bibitem{sen2019think}
R.~Sen, H.-F. Yu, I.~S. Dhillon, Think globally, act locally: A deep neural network approach to high-dimensional time series forecasting, Advances in neural information processing systems 32 (2019).

\bibitem{song2020spatial}
C.~Song, Y.~Lin, S.~Guo, H.~Wan, Spatial-temporal synchronous graph convolutional networks: A new framework for spatial-temporal network data forecasting, in: Proceedings of the AAAI Conference on Artificial Intelligence, Vol.~34, 2020, pp. 914--921.

\bibitem{wu2023interpretable}
H.~Wu, H.~Zhou, M.~Long, J.~Wang, Interpretable weather forecasting for worldwide stations with a unified deep model, Nature Machine Intelligence (2023) 1--10.

\bibitem{jeon2022gt}
J.~Jeon, J.~Kim, H.~Song, S.~Cho, N.~Park, Gt-gan: General purpose time series synthesis with generative adversarial networks, in: Advances in Neural Information Processing Systems, 2022.

\bibitem{zhang2019trafficgan}
Y.~Zhang, S.~Wang, B.~Chen, J.~Cao, Z.~Huang, Trafficgan: Network-scale deep traffic prediction with generative adversarial nets, IEEE Transactions on Intelligent Transportation Systems 22~(1) (2019) 219--230.

\bibitem{li2021fourier}
Z.~Li, N.~B. Kovachki, K.~Azizzadenesheli, B.~liu, K.~Bhattacharya, A.~Stuart, A.~Anandkumar, Fourier neural operator for parametric partial differential equations, in: International Conference on Learning Representations, 2021.

\bibitem{vaswani2017attention}
A.~Vaswani, N.~Shazeer, N.~Parmar, J.~Uszkoreit, L.~Jones, A.~N. Gomez, {\L}.~Kaiser, I.~Polosukhin, Attention is all you need, Advances in neural information processing systems 30 (2017).

\bibitem{gulrajani2017improved}
I.~Gulrajani, F.~Ahmed, M.~Arjovsky, V.~Dumoulin, A.~C. Courville, Improved training of wasserstein gans, Advances in neural information processing systems 30 (2017).

\bibitem{chen2001freeway}
C.~Chen, K.~Petty, A.~Skabardonis, P.~Varaiya, Z.~Jia, Freeway performance measurement system: mining loop detector data, Transportation Research Record 1748~(1) (2001) 96--102.

\bibitem{jagadish2014big}
H.~V. Jagadish, J.~Gehrke, A.~Labrinidis, Y.~Papakonstantinou, J.~M. Patel, R.~Ramakrishnan, C.~Shahabi, Big data and its technical challenges, Communications of the ACM 57~(7) (2014) 86--94.

\bibitem{asuncion2007uci}
A.~Asuncion, D.~Newman, Uci machine learning repository (2007).

\bibitem{chen2015ucr}
Y.~Chen, E.~Keogh, B.~Hu, N.~Begum, A.~Bagnall, A.~Mueen, G.~Batista, The ucr time series classification archive (2015).

\end{thebibliography}











\end{document}